\documentclass[10pt,twocolumn,letterpaper]{article}

\usepackage[pagenumbers]{cvpr} 

\definecolor{cvprblue}{rgb}{0.21,0.49,0.74}
\usepackage[pagebackref,breaklinks,colorlinks,allcolors=cvprblue]{hyperref}
\usepackage[ruled, vlined]{algorithm2e}
\usepackage{graphicx}
\usepackage{multicol}
\usepackage{multirow}
\usepackage{wrapfig}
\usepackage{cuted}
\usepackage{pifont}
\usepackage[accsupp]{axessibility}

\def\paperID{7222} 
\def\confName{CVPR}
\def\confYear{2025}

\title{Improving Editability in Image Generation with Layer-wise Memory}
\author{
Daneul Kim \hspace{0.5cm} Jaeah Lee\hspace{0.5cm} Jaesik Park\thanks{Corresponding author.}
\\
Seoul National University, Republic of Korea\\
\texttt{\small\{carpedkm, hayanz, jaesik.park\}@snu.ac.kr}
}

\usepackage{arydshln}
\makeatletter
\def\adl@drawiv#1#2#3{%
        \hskip.5\tabcolsep
        \xleaders#3{#2.5\@tempdimb #1{1}#2.5\@tempdimb}%
                #2\z@ plus1fil minus1fil\relax
        \hskip.5\tabcolsep}
\newcommand{\cdashlinelr}[1]{%
  \noalign{\vskip\aboverulesep
           \global\let\@dashdrawstore\adl@draw
           \global\let\adl@draw\adl@drawiv}
  \cdashline{#1}
  \noalign{\global\let\adl@draw\@dashdrawstore
           \vskip\belowrulesep}}
\makeatother

\begin{document}
\maketitle
\begin{strip}
    \centering
    \vspace{-1.5cm}
    \includegraphics[trim={1mm 2mm 1mm 1mm}, clip, width=1.0\textwidth]{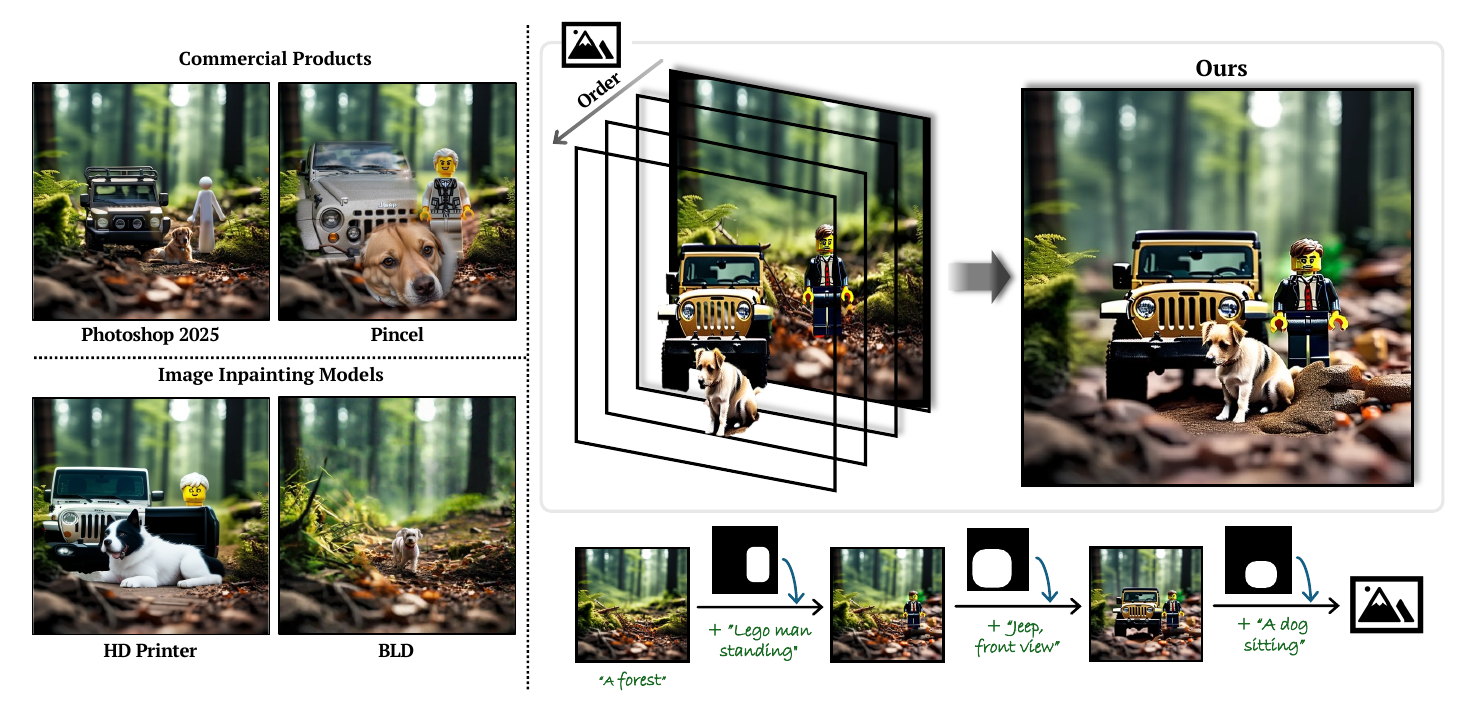}
    \vspace{-10pt}
  \captionof{figure}{\textbf{Overview.} Our framework enables the interactive generation of images with enhanced control but in a simple manner, by rough mask and prompt, through iterative scene editing. We utilize the background scene generated by our framework to edit in HD Painter~\cite{manukyan2023hd} or Blended Latent Diffusion (BLD)~\cite{avrahami2023blended} for comparison and commercial products like Photoshop~\cite{photoshop} and Pincel~\cite{pincel}.}
  \label{fig:motivation}
\end{strip}

\begin{abstract}
Most real-world image editing tasks require multiple sequential edits to achieve desired results.
Current editing approaches, primarily designed for single-object modifications, struggle with sequential editing: 
especially with maintaining previous edits along with adapting new objects naturally into the existing content.
These limitations significantly hinder complex editing scenarios where multiple objects need to be modified while preserving their contextual relationships.
We address this fundamental challenge through two key proposals: 
enabling rough mask inputs that preserve existing content while naturally integrating new elements and supporting consistent editing across multiple modifications. Our framework achieves this through layer-wise memory, which stores latent representations and prompt embeddings from previous edits. We propose Background Consistency Guidance that leverages memorized latents to maintain scene coherence and Multi-Query Disentanglement in cross-attention that ensures natural adaptation to existing content.
To evaluate our method, we present a new benchmark dataset incorporating semantic alignment metrics and interactive editing scenarios. Through comprehensive experiments, we demonstrate superior performance in iterative image editing tasks with minimal user effort, requiring only rough masks while maintaining high-quality results throughout multiple editing steps.
\end{abstract}

\section{Introduction}

Recent advances in text-to-image synthesis through powerful diffusion models like Stable Diffusion~\cite{rombach2022high, sd3}, PixArt~\cite{chen2023pixartalpha, chen2024pixartsigma} and FLUX~\cite{flux} have transformed visual content creation. 
However, users often demand multiple sequential edits to achieve desired results, iteratively refining and adding elements to their images.
Current approaches in inpainting~\cite{manukyan2023hd, avrahami2023blended} are primarily designed for single-object modifications with limited changes like colors or styles, making them inadequate for complex editing scenarios that involve multiple objects and their interactions.

Image editing approaches, while offering localized modifications, face several limitations. 
Current methods~\cite{Chen2024CATiffusion,manukyan2023hd, avrahami2023blended,smartbrush2023} struggle when modifications need to be applied sequentially, often demanding precise segmentation masks~\cite{Chen2024CATiffusion} or external modules~\cite{sam2} to maintain background integrity~\cite{manukyan2023hd}.
For example, generating \emph{``A dog sitting''} as shown in Fig.~\ref{fig:motivation} remains challenging, as it requires maintaining previous edits while ensuring the \emph{``dog''} naturally blends with the \emph{``Jeep''} and \emph{``Lego man standing''} -- a complex iterative editing scenario that current methods~\cite{manukyan2023hd,avrahami2023blended, photoshop, pincel} struggle to handle.

Layout-to-image generation offers an alternative approach through various inputs including bounding boxes~\cite{li2023gligen, xie2023boxdiff, zheng2023layoutdiffusion}, depth maps~\cite{lee2024compose, zhang2023adding}, and semantic masks~\cite{bar2023multidiffusion, lee2024streammultidiffusion}.
However, these methods require complete regeneration of the entire image for each modification, making iterative editing particularly cumbersome -- unable to maintain the surrounding background.  
While recent work has explored layered representations~\cite{ren2024move,huang2024layerdiff,zhang2023text2layer} and instance-based generation~\cite{wang2024instancediffusion}, they require complex optimization processes or additional training, limiting their practical applicability in iterative editing scenarios.

We address these limitations through two key innovations. 
First, we enable object placement aligning with the user's intention using only rough mask inputs while preserving background context. 
Second, we support consistent iterative editing across multiple modifications. 
To achieve this, we introduce the concept of \emph{mask order}, which specifies the sequence of object generation during iterative image editing.
In Fig.~\ref{fig:motivation}, we add \emph{``a lego man''}, \emph{``Jeep''} sequentially with different mask order, but they naturally adapt into the image, side by side.
If we add \emph{``A dog sitting''} with overlapping mask, then it means that we aim to put the \emph{``dog''} in front of the \emph{``Jeep''} and \emph{``a lego man''}, making the instance order of \emph{``dog''} in front of \emph{``Jeep''} and \emph{``a lego man''}.

To handle this, we incorporate three technical components: 
(1) Layer-wise memory for storing editing history, 
(2) Background Consistency Guidance (BCG) for maintaining unedited regions, 
and (3) Multi-Query Disentanglement (MQD) in cross-attention for natural object integration.
The layer-wise memory stores and manages the latent representations and prompt embeddings from previous editing steps, eliminating redundant computations typical in sequential modifications while maintaining consistency across multiple edits.
The BCG not only ensures unedited regions remain stable but also reduces computational overhead by avoiding repeated forward passes on the original image, enabling reduced editing time, while MQD enables the natural integration of new objects with existing content.

Additionally, we propose Multi-Edit Bench, a comprehensive benchmark for evaluating iterative image editing capabilities. Prior benchmarks either focus on single-turn edits~\cite{i2ebench,editbench} or layout-to-image generation~\cite{bakr2023hrs,feng2023layoutgpt,layoutbench}, failing to capture the challenges of sequential modifications. Our benchmark introduces layer-wise semantic evaluation metrics to assess both edit quality and cross-modification consistency in multi-step editing scenarios.

To summarize, our key contributions include:
\begin{itemize}  
    \item A framework for interactive mask order-based object placement editing with only rough mask inputs.
    \item A localized editing mechanism using layer-wise memory that maintains consistency across multiple edits. 
    \item Multi-query disentangled cross attention, allowing natural integration while preserving background context.
    \item Multi-Edit Bench, a comprehensive benchmark for evaluating the semantic alignment and sequential iterative editing capabilities.
\end{itemize}

\section{Related Work}
\label{sec:related_work}

\noindent\textbf{Image Inpainting.}
Image inpainting has evolved from classical patch-based~\cite{criminisi2004region,barnes2009patchmatch} and interpolation-driven~\cite{ballester2001filling,bertalmio2000image} methods to powerful deep-learning approaches~\cite{yang2017high,yi2020contextual,yu2019free} that better handle complex structures. 
Text-conditioned inpainting~\cite{yu2019free,zhang2020text} further broadened user control, and diffusion-based models~\cite{ho2020denoising}—including GLIDE~\cite{nichol2021glide}, Stable Diffusion~\cite{rombach2022high}, and Blended Diffusion~\cite{avrahami2022blended,avrahami2023blended}—significantly improved realism. 
Yet, existing works often focus on single-object edits. 
Recent attempts to relax mask precision~\cite{smartbrush2023,manukyan2023hd} or enhance quality~\cite{Chen2024CATiffusion} remain limited in multi-step scenarios, struggling to maintain consistency across multiple objects or occlusions—hence insufficient for iterative real-world editing tasks.

\noindent\textbf{Layout-to-Image Generation.}
Controllable image synthesis has recently emphasized spatial arrangement via bounding boxes, segmentation masks, or other layout cues~\cite{wang2024instancediffusion,chen2024training,zhao2019image}. 
Advances in diffusion-based layout calibration~\cite{gong2024check} and grounding~\cite{shirakawa2024noisecollage} have improved fidelity, and interactive methods~\cite{chen2024anyscene,lv2024place,nitzan2024lazy} aim to let users refine elements step by step. 
However, most such pipelines regenerate the entire image for each new edit, losing previously established context and coherence. 
By contrast, our approach incrementally updates only relevant regions through a memory-assisted process, naturally preserving edited objects.

\begin{figure*}[t]
    \centering
    \includegraphics[width=0.97\linewidth]{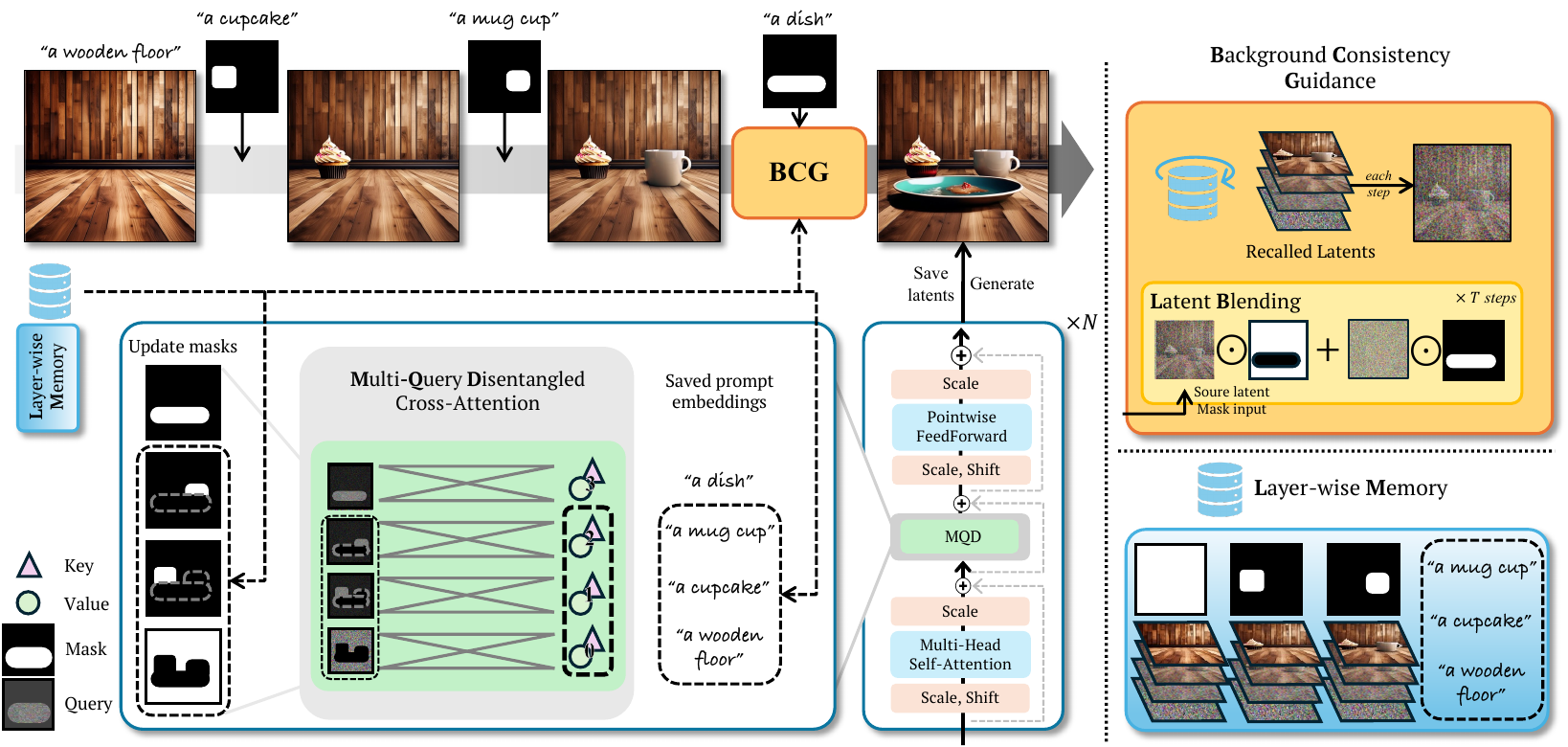}
\caption{\textbf{Overview.} (a) The left denotes an illustration of how Multi Query Disentanglement is performed in the cross-attention layer. (b) The upper right figure shows Background Consistency Guidance with recalled latents, conducting latent blending with the saved latents. (c) The right below shows the layer-wise memory, saving the previous editing steps' latents, masks, and prompt embeddings.}
    \label{fig:overview}
    \vspace{-0.2cm}
\end{figure*}

\noindent\textbf{Evaluation Benchmark.}
Most existing editing benchmarks focus on single-turn modifications~\cite{i2ebench,editbench}, while layout-centric datasets such as HRS-1k~\cite{bakr2023hrs} and NSR-1k~\cite{feng2023layoutgpt} assess scene quality or prompt alignment in a single-shot manner. Consequently, they overlook the iterative nature of real-world edits where object placements evolve over multiple steps. To address this gap, we introduce \emph{Multi-Edit Bench}, a new benchmark that evaluates multi-step editing in terms of semantic accuracy and visual alignment across the entire editing sequence. This multi-step perspective better captures practical editing workflows, where users iteratively refine scenes by adding, removing, or repositioning objects, thereby measuring a method's capacity to preserve context and maintain coherent compositions over multiple, potentially complex modifications.

\section{Method}

\subsection{Overview}
We propose a framework for iterative image editing with three key components: Layer-wise memory, background consistency guidance (BCG), and multi-query disentangled cross-attention (MQD). 
Layer-wise memory stores editing history to enable consistent object placement across sequential edits.
BCG leverages this stored information for efficient latent blending while preserving background integrity.
MQD ensures the natural object integration through disentangling the attention between queries and latents, handling complex scenarios as shown in Fig.~\ref{fig:overview} like placing \emph{``a dish''} in front of \emph{``a cupcake''} and \emph{``a mug cup''}.

Our framework's architecture, built upon transformer-based diffusion models~\cite{chen2023pixartalpha}, operates through the following systematic workflow: First, we detail the layer-wise memory mechanism (Fig.~\ref{fig:overview}(c)) in Sec.~\ref{sec:layerwise}, which stores editing history. 
Then, in Sec.~\ref{sec:bcg}, we explain how BCG leverages this stored information for efficient latent blending (Fig.~\ref{fig:overview} (b)). Sec.~\ref{sec:mqd} describes MQD's role in ensuring natural object integration through query disentanglement (Fig.~\ref{fig:overview} (a)). 
Finally, in Sec.~\ref{sec:improving}, we demonstrate how these components enable advanced editing capabilities, including the removal of occluded elements while preserving foreground integrity.

\subsection{Layer-wise Memory}\label{sec:layerwise}
Layer-wise memory enables stable background preservation while ensuring the natural integration of new objects during iterative editing with \emph{mask order}. 
By storing and managing key information from each editing step according to the specified mask order, the model can reference previous edits when generating new content while maintaining the intended spatial relationships between objects. 

Let the set of prompts for generating the background and objects be denoted as $ P_l = \{p_0, p_1, p_2, \dots\} $, where $ p_0 $ is the background generation prompt, and $ p_1, p_2, \dots $ represent the prompts for each object. Similarly, let the set of corresponding masks be denoted as $ M_l = \{m_0, m_1, m_2, \dots\} $, where each $ m_i $ defines the region of interest (RoI) for the object associated with prompt $ p_i $. Note that $m_0$ represents the case with mask of all elements being $1$.

The layer-wise memory $ L_l = \{l_0, l_1, l_2, \dots\} $ stores three key elements for each step $ i $:
$
l_i = \{\mathbf{p}_i, \{\mathbf{Z}_i^t\}_{t=1}^{T}, m_i \}
$, where:
$ \mathbf{p}_i $ denotes prompt embedding that guides the generation, $ \{\mathbf{Z}_i^t\}_{t=1}^{T} $ as denoising latents across timesteps
and $ m_i $ as a mask defining the object's RoI.

For a generation, the initial background latent is created as $ \mathbf{Z}_0 = f_\theta(p_0) $, where $ f_\theta $ is the pre-trained diffusion model. For each subsequent object $ p_i $, we initialize its latent independently: $ \mathbf{Z}_i = f_\theta(p_i, m_i) $. This independence in initialization, combined with the stored history in layer-wise memory, enables flexible generation while maintaining background consistency.

The layer-wise memory serves two key purposes:
(1) Context preservation, to store the complete editing history, enabling the model to maintain consistency with previously generated content
(2) Localized editing, enabling control over specific regions while preserving the surrounding content through mask-guided generation

By maintaining this structured history of edits, the model can effectively handle complex scenarios where multiple objects need to be integrated coherently. The stored information guides background consistency and natural object integration, as detailed in the following sections.

\begin{table}[t]
    \centering
    \footnotesize
    \caption{\textbf{Computational comparison in utilizing Background Consistency Guidance (BCG).} We utilize PixArt-$\alpha$ inpaint pipeline with vanilla latent blending and our custom-implemented pipeline with BCG for comparison. We ran the inpainting task for the single-step editing 5 times for each method. Note that we use PixArt-alpha-XL-1024 pretrained weight for the experiment.}
    \resizebox{0.75\columnwidth}{!}{
    \begin{tabular}{ccc}
    \toprule
     Method & Time (sec) & VRAM \\ 
       \cmidrule(lr){1-1} \cmidrule(lr){2-2} \cmidrule{3-3}
    Latent Blending & 4.1218 $\pm 0.0521$ & \textbf{16.89GB} \\
    \textbf{BCG} & \textbf{3.8992$\pm \textbf{0.0142}$} & 16.90GB  \\
   \bottomrule
   \vspace{-7mm}
    \end{tabular}
    }
\label{tab:bcg_timecomp}
\end{table}
\subsection{Background Consistency Guidance}
\label{sec:bcg}

Background Consistency Guidance (BCG) addresses a key challenge in iterative image editing: maintaining the integrity of previous edits while incorporating new elements efficiently. Unlike traditional inpainting approaches, BCG leverages layer-wise memory to ensure stability across multiple sequential modifications.

\vspace{2mm}
\noindent\textbf{Latent Retrieval from Layer-wise Memory:}  
For each editing step $i$, BCG retrieves information from the previous edits stored in layer-wise memory $L_l$, enabling seamless integration of new content while preserving prior modifications:
\begin{equation}
l_{i-1} = \{\mathbf{p}_{i-1}, \{\mathbf{Z}_{i-1,t}\}_{t=1}^{T}, m_{i-1}\}
\end{equation}
where $\mathbf{p}_{i-1}$ is the prompt embedding from the prior step, $\{\mathbf{Z}_{i-1,t}\}_{t=0}^{T}$ represents the stored latents from each denoising step, and $m_{i-1}$ is the previous mask. This stored latent provides context for the current edit, preserving the background information without requiring a re-computation on the original image.

\vspace{2mm}
\noindent\textbf{Ensuring Consistency:}  
When adding a new object, we update only the masked region while preserving the rest:
\begin{equation}
\mathbf{Z}_i = \mathbf{Z}_{i-1} \odot (1 - m_i) + \mathbf{Z}_i \odot m_i
\end{equation}
This selective update mechanism is crucial for iterative editing since it ensures that each new modification preserves previous edits. The element-wise multiplication $\odot$ enables control over which regions are updated, maintaining the integrity of the evolving image composition.

\vspace{2mm}
\noindent\textbf{Computational Efficiency for Iterative Editing:}  

BCG improves the efficiency of iterative image editing by reducing redundant computations in sequential modifications. 
Traditional inpainting methods~\cite{avrahami2023blended} that utilize latent blending (LB) require a forward pass on the original image for each edit, becoming costly with multiple edits.
While both approaches (\ie, LB, BCG) require denoising costs, BCG avoids repeated forward passes.
Let $\Omega = \text{FLOPs}(T, L, H, W)$ represent the base computational cost for processing an image of size $H \times W$ through $T$ denoising steps and $L$ model layers, and $C_f$ denote the forward pass cost.
The total cost for LB and BCG can be expressed as:
\begin{equation}
\text{Cost}_{\text{LB}} = C_f + \Omega.
\end{equation}
While for BCG, by skipping the forward pass:
\begin{equation}
\text{Cost}_{\text{BCG}} = \Omega,
\end{equation}
Assuming $C_f \approx r\Omega$ with ratio $r \in (0, 1)$, we can estimate:
\begin{equation}
\text{Efficiency Gain} = \frac{\text{Cost}_{\text{LB}}}{\text{Cost}_{\text{BCG}}} \approx 1 + r
\end{equation}
In Tab.~\ref{tab:bcg_timecomp}, we observe about 10\% reduction in computational time for single-step editing with BCG compared to LB. This improvement becomes more significant in scenarios requiring multiple sequential modifications, as LB requires multiple forward passes while BCG requires none.

\subsection{Multi-Query Disentangled Cross-Attention}
\label{sec:mqd}

Multi-query disentangled cross-attention (MQD) ensures the natural integration of objects across different mask orders while preserving their spatial relationships. 
For example, when generating ``a dish'' as in Fig.~\ref{fig:overview}, MQD enables the `dish' to blend naturally into the background, despite being in different mask orders, while maintaining the structure and color of the background with proper occlusion of other objects like `a mug cup' and `a cupcake' behind. 
This is achieved by disentangling attention across multiple queries and leveraging information from layer-wise memory $ L_l $.

\begin{figure*}[t]
  \centering
  \includegraphics[trim={3mm 2mm 2mm 2mm}, clip, width=0.9\textwidth]{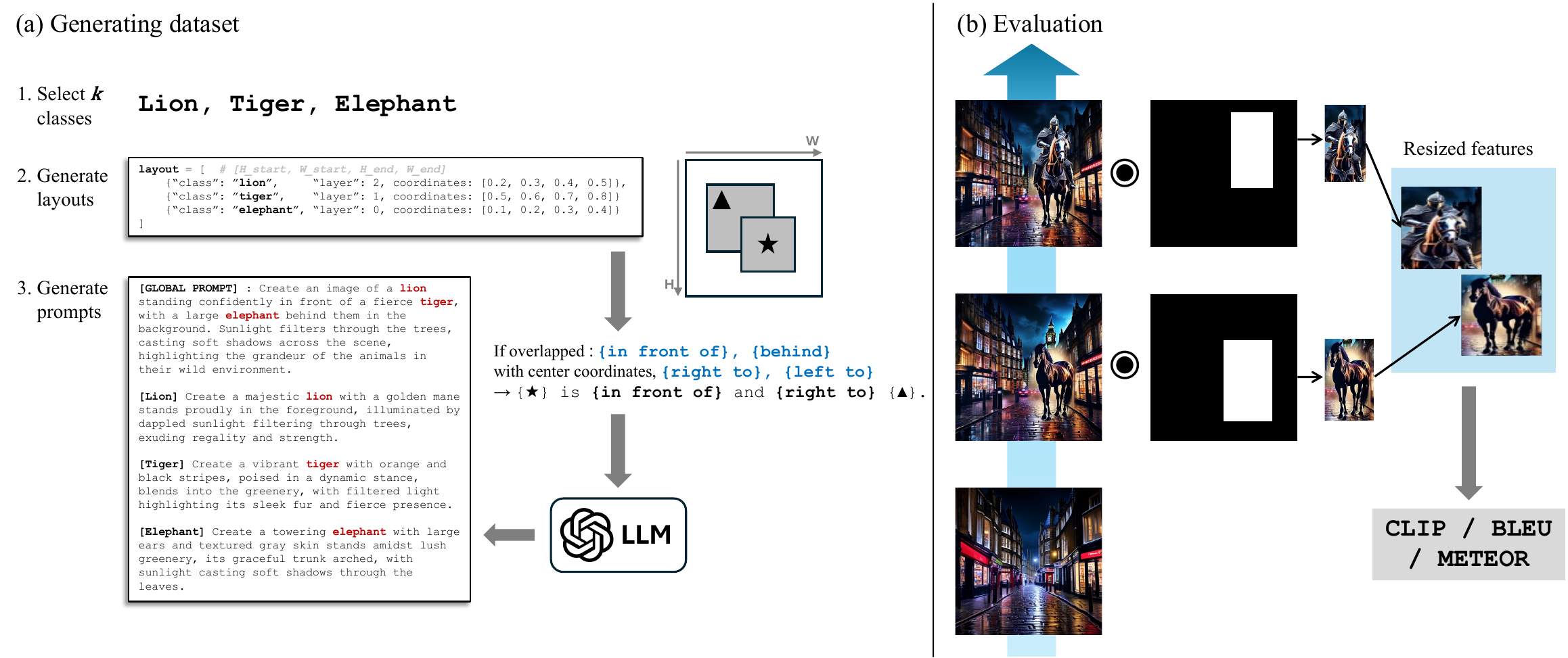}
  \vspace{-2mm}
  \caption{\textbf{Overview of our proposed Multi-Edit Benchmark for evaluation of iterative editing scenario.} (a) explains the dataset generation pipeline through GPT-4 API, and (b) explains the evaluation methodology in visual alignment using CLIP and semantic alignment using LLaVa for single-image and in a layer-wise manner.}
  \label{fig:exp_dataaset}
  \vspace{-4mm}
\end{figure*}

\subsubsection{Cross-Attention for Current Object}
For each generation step $ i $, MQD focuses attention on the current mask order's region of interest (RoI):
\begin{equation}
\mathbf{z}_i^{k,attn} = \text{CrossAttention}(\mathbf{z}_i^{k,t} \odot m_i, p_i)
\end{equation}
where $ \mathbf{z}_i^{k,t} $ is the latent for the $k$-th block at denoising step $t$, $ m_i $ defines the RoI, and $ p_i $ is the prompt embedding. This ensures focused generation within the current mask order's region. In the Fig.~\ref{fig:overview}, $m_i$ is given as a mask corresponding to the $p_i$ `a dish' with $\mathbf{z}_i^{k,t}$ corresponding to the latent of the masked region for `a dish'.

\subsubsection{Disentangled Attention for Previous Mask Orders}
To adapt the mask orders into the appropriate instance orders, MQD applies attention to non-overlapping regions from previous steps:
\begin{equation}
\mathbf{z}_i^{k,attn} = \displaystyle\bigcup_{j=0}^{i-1} \text{CrossAttention}(\mathbf{z}_i^{k,t} \odot (m_j - \displaystyle\Sigma_{l=j+1}^{i} m_l), p_j)
\end{equation}
This operation ensures that objects from earlier mask orders, (\ie, previous masks for `a mug cup' and `a cupcake' on top of `a wooden floor' behind `a dish' altogether) remain coherent while allowing natural adaptation where needed. The term $ m_j - \displaystyle\Sigma_{l=j+1}^{i} m_l $ controls which regions from previous mask orders should influence the current generation, minimally affecting the \emph{``mug cup''} and \emph{``cupcake''} even if we put the \emph{``dish''} in front of them.
Please note that for background prompt $p_0$: \emph{``wooden floor''}, we inverse the whole union of masks and do disentangled cross attention with background prompt $p_0$ as in Fig.~\ref{fig:overview} (a).

\subsubsection{Final Latent Update}
The final update combines information across all mask orders:
\begin{equation}
\mathbf{z}_i^{merge} = \mathbf{z}_i^{attn} \bigcup \sum_{j=0}^{i-1} \mathbf{z}_j^{attn}
\end{equation}
After a feedforward layer update and $K$ transformer blocks, we apply BCG:
\begin{equation}
\mathbf{Z}_i^t = \mathbf{z}_i^{K,t} \odot m_i + \mathbf{Z}_{i-1}^t \odot (1 - m_i)
\end{equation}

This approach enables objects to adapt naturally to mask orders—for example, placing \emph{``a dog''} in front of \emph{``Jeep''} and \emph{``Lego man''} (Fig.~\ref{fig:motivation}), or maintaining `a cupcake' and `a mug cup' despite inserting \emph{``a dish''} in front (Fig.~\ref{fig:overview})—while preserving scene coherence. 
MQD balances content preservation with the seamless integration of new elements.

\subsection{Improving Editability}\label{sec:improving}
While BCG and MQD with layer-wise memory enhance multiple editing capabilities in the PixArt-$\alpha$ model, we further extend our framework to support object deletion, particularly for overlapped objects behind foreground elements. We achieve this through a novel application of our MQD and BCG components.

To delete an object added at editing step $i-1$, we leverage the stored latent representations from both step $i-2$ and the current step $i$: $\{\mathbf{Z}_{i-2}^t\}_{t=1}^T$ and $\{\mathbf{Z}_{i}^t\}$. Starting from an intermediate denoising step $\tau$, we blend these latents as:
\begin{equation}
    \mathbf{Z}_{erase}^\tau = m_i \odot \mathbf{Z}_{i}^\tau + (1 - m_i) \odot \mathbf{Z}_{i-2}^\tau
\end{equation}

Our deletion process operates in two phases: from step $\tau$ to $\frac{\tau}{2}$, we perform latent blending following Eq. (10) with $\mathbf{Z}_{i-2}^t$. After step $\frac{\tau}{2}$, we switch to vanilla denoising to ensure a smooth transition. Crucially, we process $\mathbf{Z}_{erase}$ through MQD without the $i-1$ step's mask or prompt embedding, effectively removing the target object's influence.

This approach offers two key advantages. First, by leveraging BCG to start from step $\tau$ rather than T, we achieve 60\% faster editing while preserving background consistency, as we use $\tau=8$ out of a total of 20 steps. 
Second, MQD's application from step $\tau$ maintains foreground object integrity while effectively removing background elements, eliminating the need for precise object masks in deletion.

\begin{table*}[t]
    \centering
    \caption{\textbf{Quantitative results on our proposed benchmark.} Our method shows higher semantic and visual alignment than other baselines, including layout-to-image synthesis and image editing frameworks. Ordering$\dagger$ denotes the adaptation of the framework adequate for iterative image generation. 
    }
    \vspace{-5pt}
    \resizebox{0.9\textwidth}{!}{
    \begin{tabular}{@{}lcccccc@{}}
    \toprule
     \multirow{2}{*}{Type} & \multirow{2}{*}{Method} & \multirow{2}{*}{Resolution} & \multicolumn{2}{c}{Semantic Alignment} & Visual Alignment \\ 
      \cmidrule(lr){4-5} \cmidrule(lr){6-6} 
      & & & BLEU-2/3/4$\uparrow$ & METEOR$\uparrow$ & CLIP\textsubscript{crop}$\uparrow$ \\
    \midrule
    \multirow{3}{*}{2D Layout-to-Image} 
    & LayoutGuidance~\cite{chen2024training} & $512\times512$ & 36.44 / 26.13 / 18.85 & 0.1361 &  62.92 \\
    & NoiseCollage~\cite{shirakawa2024noisecollage} & $512\times512$ & 55.75 / 42.43 / 32.96 & 0.1402 & 64.01 \\
    & NoiseCollage + ordering$\dagger$ & $512\times512$  & 59.98 / 43.76 / 32.24 & 0.1464 &  64.10 \\
    \midrule
    \multirow{1}{*}{3D Layout-to-Image}&LooseControl~\cite{loosecontrol} & $512\times512$& 63.30 / 46.24 / 34.15 & 0.1373 & 63.13 \\
    \midrule
    \multirow{5.5}{*}{Image Editing} 
    & BLD~\cite{avrahami2023blended} & $1024\times1024$ & 55.30 / 40.38 / 29.58 & 0.1480 &  62.40 \\
    & HD-Painter~\cite{manukyan2023hd} & $1024\times1024$ & 63.29 / 47.63 / 36.28 & 0.1484 & 64.09 \\
    & SD3-Inpaint~\cite{sd3} & $1024\times1024$& 29.90 / 21.64 / 15.78 & 0.1445 & 63.98 \\
    \cmidrule{2-7}
    & Ours & $512\times512$ & 61.19 / 45.04 / 34.06 & 0.1465 & 64.28\\
    \cdashlinelr{2-7}
    & Ours & $1024\times1024$ & \textbf{64.99 / 47.69 / 36.59} & \textbf{0.1513} & \textbf{64.29}  \\
    \bottomrule
    \end{tabular}
    }
    \vspace{-5pt}
    \label{tab:quanti}
\end{table*}

\section{Multi-Edit Benchmark}

Prior benchmarks for image editing~\cite{i2ebench,editbench} are limited to evaluate only single-step editing or limited to instruct-based edit~\cite{seeddataedit}, and layout-guided image generation benchmarks are limited to evaluate whether the specific object is generated or not in the bounding box or evaluate only the flat spatial arrangements.~\cite{bakr2023hrs,feng2023layoutgpt,layoutbench}. 
Therefore, we aim to tackle the issue of evaluating the semantic alignment of the generated images by assessing whether the object generated is aligned with the caption that is used for sequential editing.
Also, to handle the mask order, we propose to evaluate the generated image in a layer-wise manner aligning with mask orders.
Specifically, to evaluate interactive generation and editing, we need to evaluate whether each layer shows the desired object within the given mask.

We generate the dataset by selecting classes from ImageNet-1K and arranging them in layered compositions with varying degrees of occlusion and spatial relationships, ensuring mask order is a key factor in the generation process as in the Fig.~\ref{fig:exp_dataaset} (a). 
We utilize GPT-4 API to assist in selecting object classes (\eg, `Lion', `Tiger', `Elephant') that naturally fit together, creating realistic compositions, and generating template-based captions for both the global scene and individual object layers to integrate spatial relations into prompt and utilize GPT to make prompt richer.

To evaluate the spatial and semantic alignment between generated images and prompts, we crop each object layer using its corresponding mask and evaluate it individually as in Fig.~\ref{fig:exp_dataaset} (b). 
Both cropped and final images are resized to 224x224 for consistency across evaluations. 
We employ LLaVa~\cite{liu2023llava} to generate captions and compute alignment metrics such as BLEU~\cite{Papineni2002BleuAM}, METEOR~\cite{Banerjee2005METEORAA}, ensuring a robust evaluation of both spatial arrangement and semantic accuracy.
Note that we average the result from all the layers of each image to evaluate.

Also, to check each layer's visual alignment of the generated object, we crop the image for each iteration and calculate the CLIP score following other layout-to-image models~\cite{feng2023layoutgpt, bakr2023hrs, shirakawa2024noisecollage} as individual editing step is similar to layout-to-image generation.
We discuss and provide detailed descriptions of the dataset creation process, evaluation metrics, and the resulting benchmark in the supplement.

\section{Experiment}

\subsection{Implementation Details}

\noindent\textbf{Baselines.}
We evaluate our method against state-of-the-art baselines in both image editing and layout-to-image generation. 
For image editing, we compare with HD-Painter~\cite{manukyan2023hd}, Blended Latent Diffusion~\cite{avrahami2023blended} with SD-XL~\cite{podell2023sdxl}, and SD3-ControlNet-Inpaint~\cite{sd3}. As sequential editing shares similarities with layout-to-image generation, we also include train-free layout models: NoiseCollage~\cite{shirakawa2024noisecollage} and Layout-Guidance~\cite{chen2024training}, and 3D-lifted model: LooseControl~\cite{loosecontrol}.

Additionally, we adapt NoiseCollage to handle sequential mask inputs (denoted as NoiseCollage + ordering) to explore the potential of converting layout-to-image models for sequential editing tasks. 
Following our train-free approach using the PixArt-$\alpha$ foundation model~\cite{chen2023pixartalpha}, all selected baselines operate without additional training.

\noindent\textbf{Implementation Details.}
We leverage PixArt-$\alpha$~\cite{chen2023pixartalpha}, the variant of Diffusion Transformer (DiT) for image editing framework.
We utilize a $1024 \times 1024$ pretrained model with DPM-Solver with DPMS guidance of 7.5 and a total denoising step of 20.
Also, for a fair comparison with models that generate $512 \times 512$ resolution~\cite{shirakawa2024noisecollage, chen2024training}, we compare quantitatively by generating $512 \times 512$ images with the $512\times512$ pretrained PixArt-$\alpha$ model.

\subsection{Quantitative Result}
We present the quantitative evaluation results on our proposed Multi-Edit Benchmark in Tab.~\ref{tab:quanti}. 
Our method consistently outperforms both image editing and layout-to-image baselines across key metrics.
Among editing baselines, we demonstrate superior performance compared to Blended Latent Diffusion (BLD)~\cite{avrahami2023blended} with SD-XL, HD-Painter~\cite{manukyan2023hd}, and SD3-ControlNet-Inpaint~\cite{sd3}.
Notably, the significant performance gap with SD3-ControlNet-Inpaint can be attributed to the suboptimal property of utilizing ControlNet-Inpainting in multi-step editing.

Our framework shows particular strength in handling sequential editing challenges, demonstrated by consistent improvements across all metrics. 
This validates our approach's effectiveness in maintaining background consistency while naturally integrating new objects - a key challenge in sequential editing tasks.

The performance gap grows further relative to 2D layout-to-image baselines, which often struggle with overlapping or cascaded placements. 
In particular, our method outperforms NoiseCollage by 4--5\%p and surpasses LayoutGuidance by over 15\%p in BLEU and outperforms in all other metrics.
Against the 3D layout-to-image method that leverages pseudo depth maps, LooseControl~\cite{loosecontrol} shows improved BLEU compared to the 2D baselines, but ours outperforms on all metrics.
(See supplement for comparison with attribute editing on LooseControl.)

While we adapt NoiseCollage~\cite{shirakawa2024noisecollage} for iterative image editing (+ ordering), NoiseCollage + ordering achieves better visual alignment scores, indicating the potential viability of adapted layout-to-image approaches. 
However, its limited improvement across semantic alignment metrics suggests that simple adaptations are insufficient for complex sequential editing scenarios. 
This underscores the need for specialized approaches like ours that can effectively handle the unique challenges of iterative image editing.

\begin{figure}[t]
  \centering
  \includegraphics[clip, width=0.85\columnwidth]{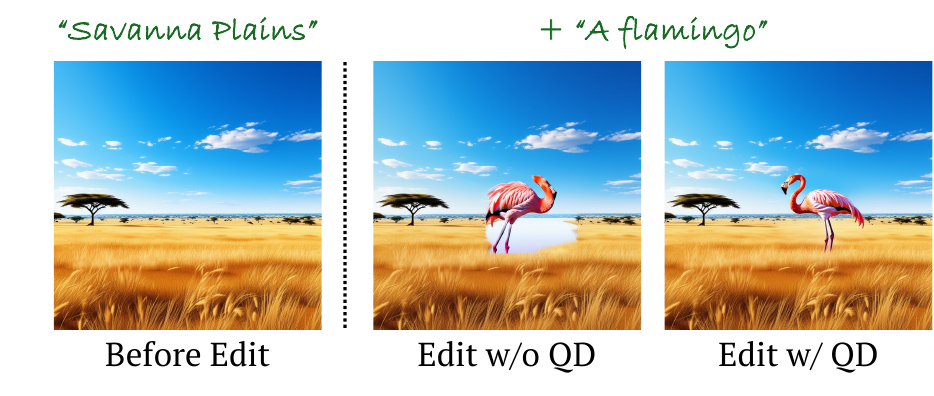}
  \vspace{-8pt}
  \caption{\textbf{Qualitative comparison on the effect of Query Disentanglement (QD).}}
  \label{fig:abl_qd}
  \vspace{-4mm}
\end{figure}
\subsection{Ablation Study}

\begin{table}[t!]
\centering
\caption{\textbf{Ablation study of our proposed method}. Results are reported on both semantic and visual alignment metrics. We denote vanilla PixArt-$\alpha$ inpainting model as \textbf{Baseline}. Also we denote \textbf{QD} as Query-Disentanglement without memory, only with the current object and background.}
\vspace{-2mm}
\resizebox{0.9\columnwidth}{!}{
\begin{tabular}{@{}lccc@{}}
    \toprule
         \multirow{2}{*}{Method} & \multicolumn{2}{c}{Semantic Align.} & Visual Align.\\
         & BLEU-2/3/4$\uparrow$& METEOR$\uparrow$ &  CLIP$_c$$\uparrow$ \\
         \midrule
         Baseline & 56.29 / 42.04 / 33.06 & \textbf{0.1586} & 64.05 \\
         ~+BCG    &  60.74 / 46.27 / 35.20& 0.1585 & 64.10 \\
         ~~+QD &  62.68 / 46.42 / 35.03 & 0.1530 &  63.99\\
         \midrule
         \textbf{Ours}     & \textbf{64.99 / 47.69 / 36.59} & 0.1513 & \textbf{64.29} \\
    \bottomrule
\vspace{-20pt}
\end{tabular}
}  
\label{tab:abl}
\end{table}

We present an ablation of our key components (Tab.~\ref{tab:abl}), starting from PixArt-$\alpha$~\cite{chen2023pixartalpha}. 
Background consistency guidance (BCG) reuses latents for faster editing (Tab.~\ref{tab:bcg_timecomp}), improving BLEU and CLIP while preserving METEOR. 
Next, Query-Disentanglement (QD), inspired by NoiseCollage~\cite{shirakawa2024noisecollage}, preserves the existing background rather than fully covering the mask, raising BLEU but slightly reducing CLIP. 
Direct latent blending instead inflates CLIP by overfilling the mask, lacking context (Fig.~\ref{fig:abl_qd}). 
Finally, extending QD to multi-query disentanglement (MQD) with layer-wise memory integrates new objects into all prior prompts, achieving higher BLEU-2/3/4 and CLIP scores.

\begin{figure}[t]
  \centering
  \includegraphics[clip, width=0.97\columnwidth]{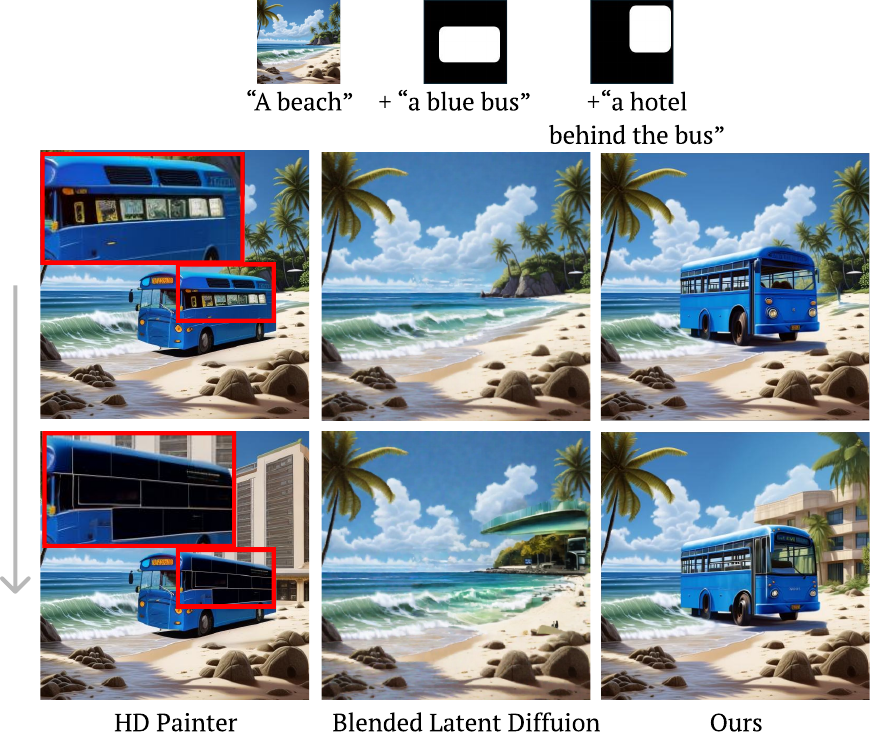}
  \vspace{-8pt}
  \caption{\textbf{Comparison in image editing capability with latest image editing models.}~\cite{manukyan2023hd, avrahami2023blended} Note that the initial image is generated by our framework, which is equivalent to PixArt-$\alpha$~\cite{chen2023pixartalpha} with no mask input.}
  \label{fig:editing}
  \vspace{-5mm}
\end{figure}

\subsection{Qualitative Result}\label{sec:quali}
\noindent\textbf{Comparison with Editing Approaches under Interactive Scenario.}
In Fig.~\ref{fig:editing}, we demonstrate our method's effectiveness in iterative editing compared to state-of-the-art single-step editing approaches~\cite{avrahami2022blended, manukyan2023hd}. 
Starting with a background of \textit{``A beach''} generated by vanilla PixArt-$\alpha$~\cite{chen2023pixartalpha}, we sequentially place a bus and a hotel. 
Our method successfully renders the hotel behind the bus, accurately interpreting both the prompt \textit{``A beach''} and the \emph{mask order} through MQD to achieve the intended spatial arrangement.

In contrast, Blended Latent Diffusion~\cite{avrahami2023blended} struggles with the initial placement of \textit{``bus''} in the complex background, while HD-Painter~\cite{manukyan2023hd} shows inconsistencies in the appearance of \textit{``bus''} (highlighted in red in Fig.~\ref{fig:editing}). 
Our method maintains background consistency while naturally integrating new objects into the scene.

\begin{figure}[t]
    \centering
    \includegraphics[trim={1mm 1mm 1mm 1mm}, clip, width=0.92\columnwidth]{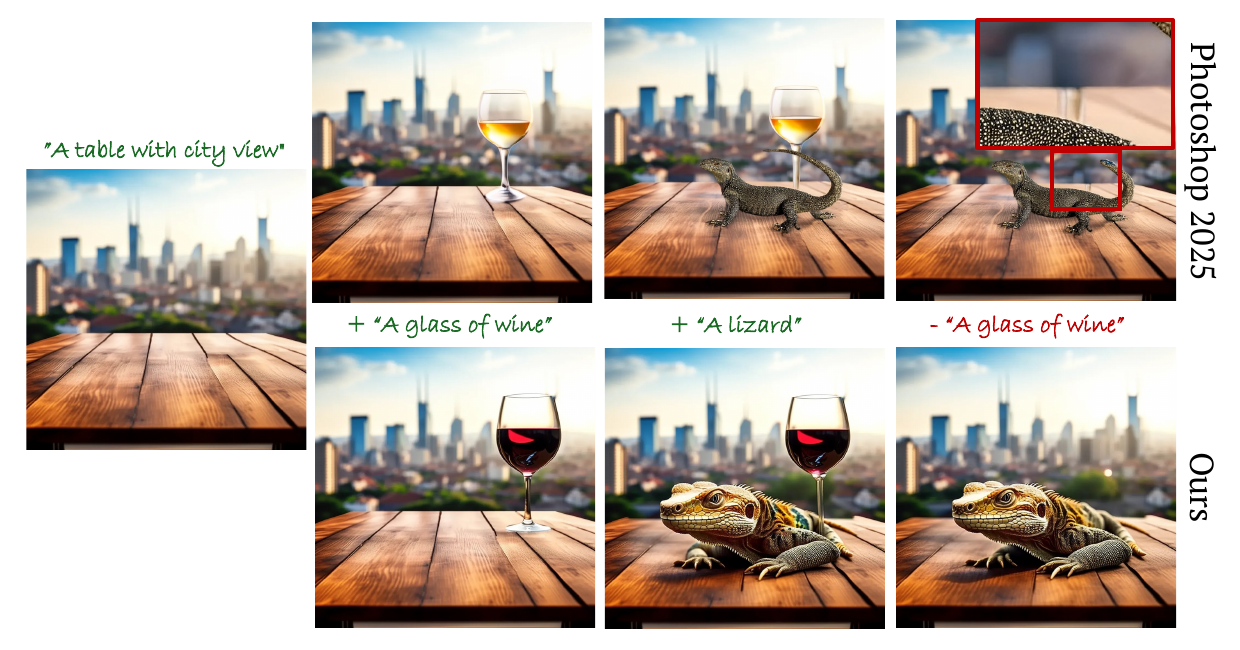}
    \vspace{-7pt}
    \caption{\textbf{Improved editability of image through Background Consistency Guidance and Multi-Query Disentangled cross attention.} Through recycling the previous step's latents, we can remove the object that is behind the foreground object, enabling enhanced editability of the image.}
    \label{fig:improved_edit}
    \vspace{-1.7mm}
\end{figure}

\noindent\textbf{Improved Editability.} 
Fig.~\ref{fig:improved_edit} compares our method with Photoshop 2025~\cite{photoshop}. 
Placing \textit{``a glass of wine''} or \textit{``a lizard''} is straightforward, but removing \textit{``a glass of wine''} while preserving \textit{``a lizard''} remains challenging for Photoshop’s generative fill, which requires precise brush strokes and often leaves artifacts (red box). 
Our approach simplifies this by leveraging existing masks and prior-step latents to naturally inpaint the background while applying MQD during partial denoising to preserve the lizard's structure.

\begin{figure}[t]
  \centering
  \includegraphics[clip, width=0.95\columnwidth]{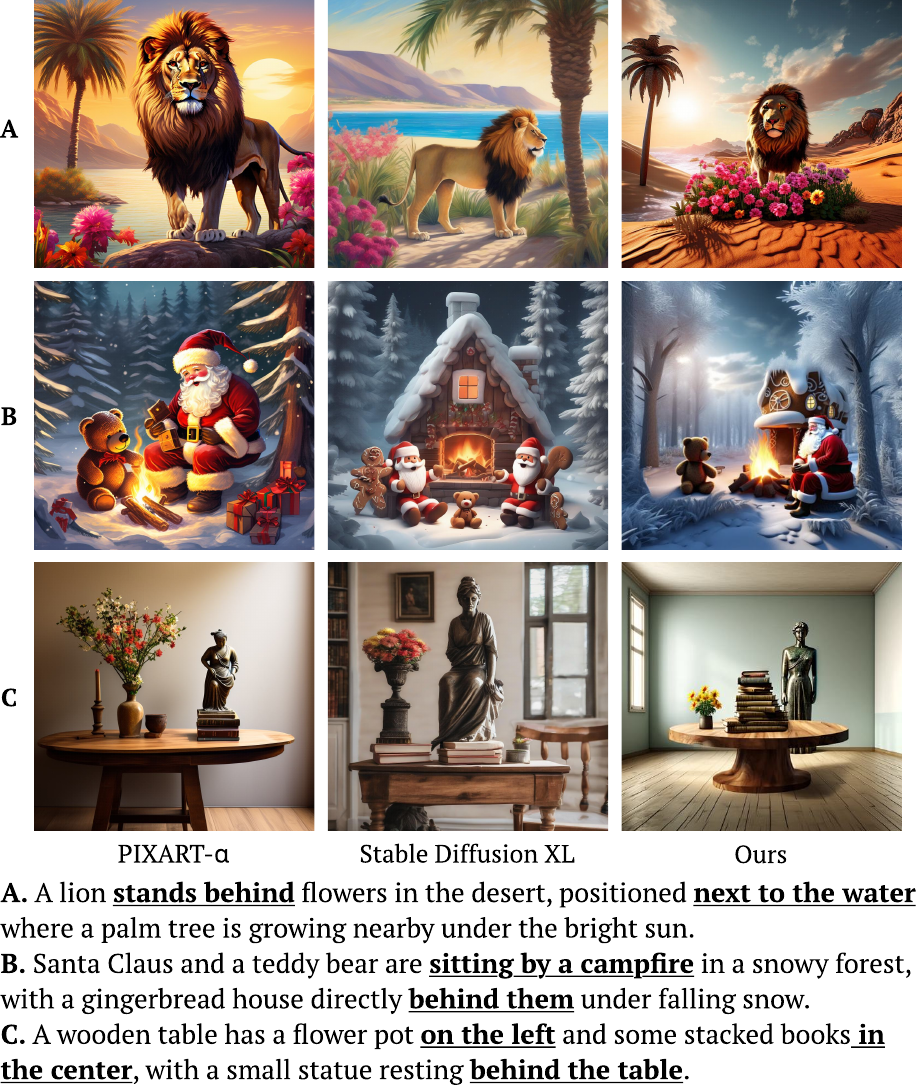}
  \vspace{-2mm}
  \caption{\textbf{Comparison in interactive scenarios with existing T2I generative models.} Stable Diffusion XL~\cite{podell2023sdxl} and PixArt-$\alpha$~\cite{chen2023pixartalpha} use text input only.}
  \label{fig:interactive_t2i}
  \vspace{-4mm}
\end{figure}

\noindent\textbf{Comparison with Existing T2I models.}
Fig.~\ref{fig:interactive_t2i} showcases our method's ability to handle complex multi-object scenarios that challenge conventional text-to-image models. 
For instance, in row A, only our framework successfully generates the intended composition of a lion behind a flower.
Row B and C further demonstrate the challenge for existing T2I approaches to handle complex spatial arrangements, while illustrating our method's capability.

\begin{table}[t]
    \centering
    \footnotesize
    \caption{\textbf{Human preference study with recent literature on image editing on our benchmark.} We utilize HD-Painter~\cite{manukyan2023hd} and Blended Latent Diffusion with SD-XL~\cite{podell2023sdxl} to compare under iterative editing scenario and compare the preference on a 5-point Likert scale.}
    \vspace{-2mm}
    \footnotesize
    \resizebox{0.85\columnwidth}{!}{
    \begin{tabular}{@{}cccc@{}}
    \toprule
     \multirow{2}{*}{Method} & \multirow{2}{*}{\shortstack{Background\\Consistency}} & \multirow{2}{*}{\shortstack{Natural\\Adaptation}} & \multirow{2}{*}{\shortstack{Text-scene\\Alignment}} \\ 
     &\\
       \cmidrule(lr){1-1} \cmidrule(lr){2-4} 
    HD-Painter~\cite{manukyan2023hd}& 3.71 & 2.81 & 3.08 \\
     BLD~\cite{avrahami2023blended}& 3.43 & 2.24 & 2.04 \\    
     \midrule
     Ours & \textbf{4.59} & \textbf{4.28} & \textbf{4.49} \\
   \bottomrule
    \end{tabular}
    }
\label{tab:human_benchmark}
\vspace{-4mm}
\end{table}

\subsection{Human Preference Study}\label{quantitative}

We conduct a comprehensive human evaluation study with 50 participants to assess the quality of our method compared to the latest editing models~\cite{manukyan2023hd, avrahami2023blended} on our benchmark. 
The evaluation focuses on three aspects: background consistency, natural adaptation, and text-scene alignment, rated on a 5-point Likert scale.

Our method outperforms single-step editing approaches~\cite{avrahami2023blended, manukyan2023hd} across all three categories. 
These results demonstrate that traditional editing models, while effective for single-step modifications, are insufficient for sequential editing.
This underscores the necessity of specialized approaches like ours for handling sequential iterative editing scenarios. 
Additional human evaluation results comparing our method with text-to-image models~\cite{podell2023sdxl, chen2023pixartalpha} are provided in the supplementary materials.

\section{Conclusion}

We propose a novel framework for sequential image editing, which poses challenges in maintaining background consistency and seamless integration of new objects into the scenes.
Our method includes three key components: \textit{Layer-wise Memory} to preserve previous content, \textit{Background Consistency Guidance} to keep the background stable with faster editing, and \textit{Multi-Query Disentangled Cross-Attention} for natural object adaptation. 
Experiments on our proposed benchmark dataset show that our approach outperforms state-of-the-art methods in semantic and visual alignment, making it ideal for complex, iterative image editing scenarios. 
This framework paves the way for future work in improved interactive editing.

\vspace{-2mm}
\paragraph{Limitation and Future Work.}
Since our approach utilizes image editing, generating multiple objects takes longer, depending on the number of edits.
Also, utilizing layer-wise memory requires additional memory costs.
We plan to make it more efficient for faster editing in the future.

\vspace{-2mm}
\paragraph{Acknowledgement.}
This work was supported by IITP grant (RS-2021-II211343: AI Graduate School Program at Seoul National University (5\%) and RS-2024-00509257: Global AI Frontier Lab (30\%)) and NRF grant (No.RS-2024-00405857 (65\%)) funded by the Korea government (MSIT).

{
    \small
    \bibliographystyle{ieeenat_fullname}
    \bibliography{main}
}

\appendix
\maketitlesupplementary
\begin{strip}
    \centering
    \includegraphics[trim={0mm 0mm 0mm 0mm}, clip, width=0.93\textwidth]{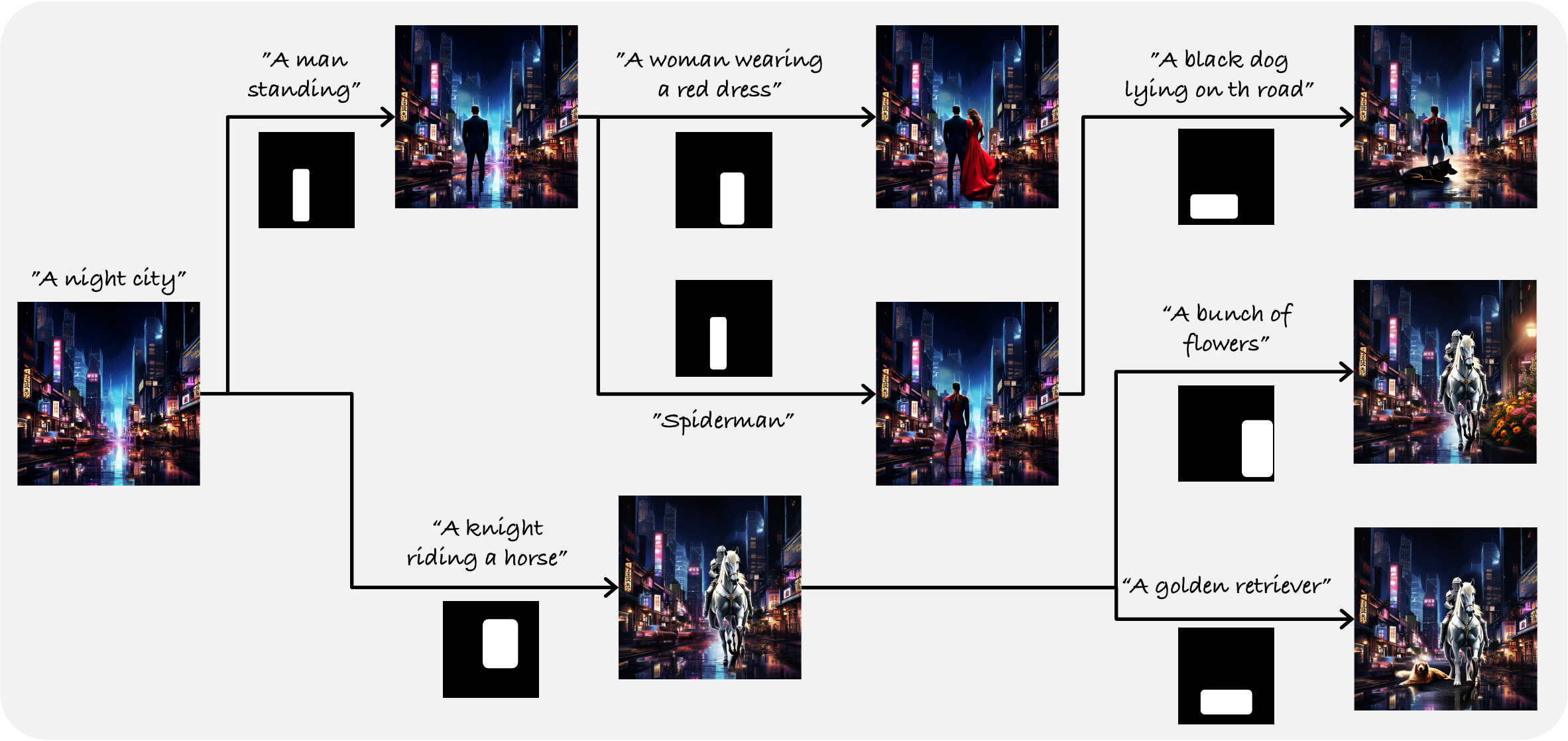}
    \captionof{figure}{\textbf{Overview of interactive image generation under various scenarios.} Our approach can easily generate diverse images by editing in different ways.}
    \label{fig:appen_interactive}
\end{strip}
\section{Implementation Details}

We provide comprehensive implementation details of our framework and baseline methods used for comparison. This section covers the technical specifications of baseline implementations, our interactive editing process, and the detailed algorithmic workflow.

\vspace{-3mm}
\paragraph{Implementation Details of Baselines.}
We compare our method against three recent image inpainting approaches: Blended Latent Diffusion (BLD)~\cite{avrahami2023blended}, HD-Painter~\cite{manukyan2023hd} and Stable Diffusion 3 (SD3)~\cite{sd3}. 
For BLD, we utilize SD-XL~\cite{podell2023sdxl} as the base model with a DDIM scheduler configured for 50 denoising steps. 

For HD-Painter, we enhance the baseline by employing DreamShaper-v8 as the pretrained weight instead of the original SD 1.5 or 2.1, ensuring better output quality.
To maintain consistent comparison, we match the resolution with our PixArt-$\alpha$ implementation using HD-Painter's built-in upscaler. 
The framework operates with a DDIM scheduler over 50 denoising steps and employs classifier-free guidance of 7.5, adhering to the original configuration.

For SD3, we use ControlNet~\cite{zhang2023adding} Inpainting version of SD3.
We use a guidance scale of 7.0 with a ControlNet scale of 0.95, with 28 inference steps, which is the original setting.

\begin{algorithm*}[ht]
\small
\caption{Layer-wise Memory with Background Consistency Guidance (BCG) and Multi-query Disentangled Cross-Attention (MQD)}
\label{alg:mqd_bcg}
\SetKwProg{Fn}{Function}{:}{\KwRet}
\SetKwInOut{Given}{Given}
\SetKw{KwInit}{Initialize}
\SetKw{KwRet}{Return}
\Given{Prompts $P_l = \{p_0, p_1, \dots, p_N\}$, Masks $M_l = \{m_0, m_1, m_2, \dots, m_N\}$, Pre-trained diffusion model $f_\theta$, Diffusion steps $T$, Number of DiT blocks $K$}
\SetNlSty{textbf}{}{:}

\textbf{Initialization:}\\
    Initialize model parameters $\theta$;\\
    Generate background latent $\mathbf{Z}_0 = f_\theta(p_0)$; \hfill \# Generate background\\
    Store $\mathbf{Z}_0$, $p_0$, $m_0$ in memory; \hfill \# Store initial background

\For{$i = 1$ \textbf{to} $N$}{
    Retrieve $\mathbf{Z}_{i-1} = \{\mathbf{Z}_{i-1}^t\}_{t=0}^T$, $p_{i-1}$, $m_{i-1}$ from memory; \hfill \# Recall previous latent
    
    Initialize Latent $\mathbf{z}_{i}^{0,T} \sim \mathcal{N}(0, I)$; \\
    \For(\hfill \# Loop over diffusion steps ){$t = T$ \textbf{to} $0$}{
    
        \For(\hfill \# Perform MQD within each DiT block){$k = 1$ \textbf{to} $K$}{ 
        
            $\mathbf{z}_{i}^{k,t} = \text{SelfAttention}(\mathbf{z}_i^{k-1,t})$;
            
            $\mathbf{z}_i^{k,attn} = \text{CrossAttention}(\mathbf{z}_i^{k,t} \odot m_i, p_i)$; \hfill \# MQD for current object
            
            \For{$j = i-1$ \textbf{to} $0$}{
                Retrieve $p_j$, $m_j$ from memory; \hfill \# Recall previous prompt embedding and mask
                
                Update $\mathbf{z}_i^{k, attn}=\text{CrossAttention}(\mathbf{z}_{i}^{k,t} \odot (m_j - \sum_{l=j+1}^{i} m_l), p_j)$; \hfill \# MQD for previous objs
            }
            Merge $\mathbf{z}_i^{merge} = \mathbf{z}_i^{attn} + \sum_{j=1}^{i-1} \mathbf{z}_j^{attn}$; \hfill \# Merge attention results
            
            $\mathbf{z}_i^{k,t} = \text{FeedForward}(\mathbf{z_i^{merge}})$
        }
        Update latent $\mathbf{Z_i^t} = \mathbf{z}_i^K \odot m_{i} + \mathbf{Z}_{i-1}^t \odot (1 - m_{i-1})$; \hfill \# Apply BCG
        
        Store $\mathbf{Z}_i^t$ in memory after final block for step $t$; \hfill \# Store final latent for each step
    }
    Store $p_i$, $m_i$ in memory after denoising; \hfill \# Store prompt embedding and mask
}

\textbf{Final Image Generation:}\\
Decode final latent $\mathbf{Z}_N$ into $\text{Image}_{\text{final}} = \text{Decoder}(\mathbf{Z}_N)$; \hfill \# Decode final latent 

\KwRet $\text{Image}_{\text{final}}$;
\end{algorithm*}
\paragraph{Interactive Editing Process.}
Our framework enables iterative image editing through a sequence of mask-guided modifications. Our framework processes each edit through three primary components: (1) Layer-wise Memory, (2) Background Consistency Guidance (BCG), and (3) Multi-Query Disentangled Cross-attention (MQD).

The Layer-wise Memory component maintains a comprehensive record of the editing history, storing latent representations, prompt embeddings, and mask information for each modification. 
This storage system enables retrieval of previous states while ensuring consistency across multiple edits. 
BCG leverages this stored information to maintain background integrity, implementing selective latent blending based on mask regions while minimizing the computational overhead of repetitive forward passes.

MQD handles the integration of new elements by processing edited regions and background content separately. 
This separation ensures the natural adaptation of new objects while preserving existing spatial relationships and background details, enabling the natural adaptation of diverse foreground objects into the background as presented in Fig.~\ref{fig:appen_interactive}.
\emph{``A man standing''} or \emph{``A knight riding a horse''} is naturally blended into \emph{``A night city''}, and when a user adds \emph{``A woman wearing a red dress''} or \emph{``A golden retriever''}, a diverse result is achieved, meeting the user's need.

\paragraph{Workflow Details.}
Algorithm~\ref{alg:mqd_bcg} presents our complete editing pipeline, which operates through four principal stages.
The process starts with initialization, where we first generate a background latent $\mathbf{Z}_0$ from the initial prompt $p_0$ and store it in layer-wise memory.
During iterative editing, we retrieve previous states ($\mathbf{Z}_{i-1}$, $p_{i-1}$, $m_{i-1}$) and process them through $T$ diffusion steps, applying $K$ DiT blocks with MQD and BCG.

The cross-attention separately processes edited regions and background content, ensuring coherent integration of new elements while preserving existing content.
BCG then blends to update the latents with \textit{retrieved latents} and stores results in layer-wise memory, maintaining a complete edit history for future modifications.
This proposed pipeline ensures robust background preservation while enabling natural object integration through the coordinated operation of our key components.

Our framework maintains editing to be coherent by leveraging MQD to disentangle cross-attention between edited regions, previously edited content and background, ensuring each modification integrates naturally with the existing scene while preserving intended spatial relationships. 
This approach enables seamless integration of new elements while maintaining the overall compositional integrity and spatial context of the image.

\begin{figure}[t]
    \centering
    \includegraphics[width=0.99\columnwidth]{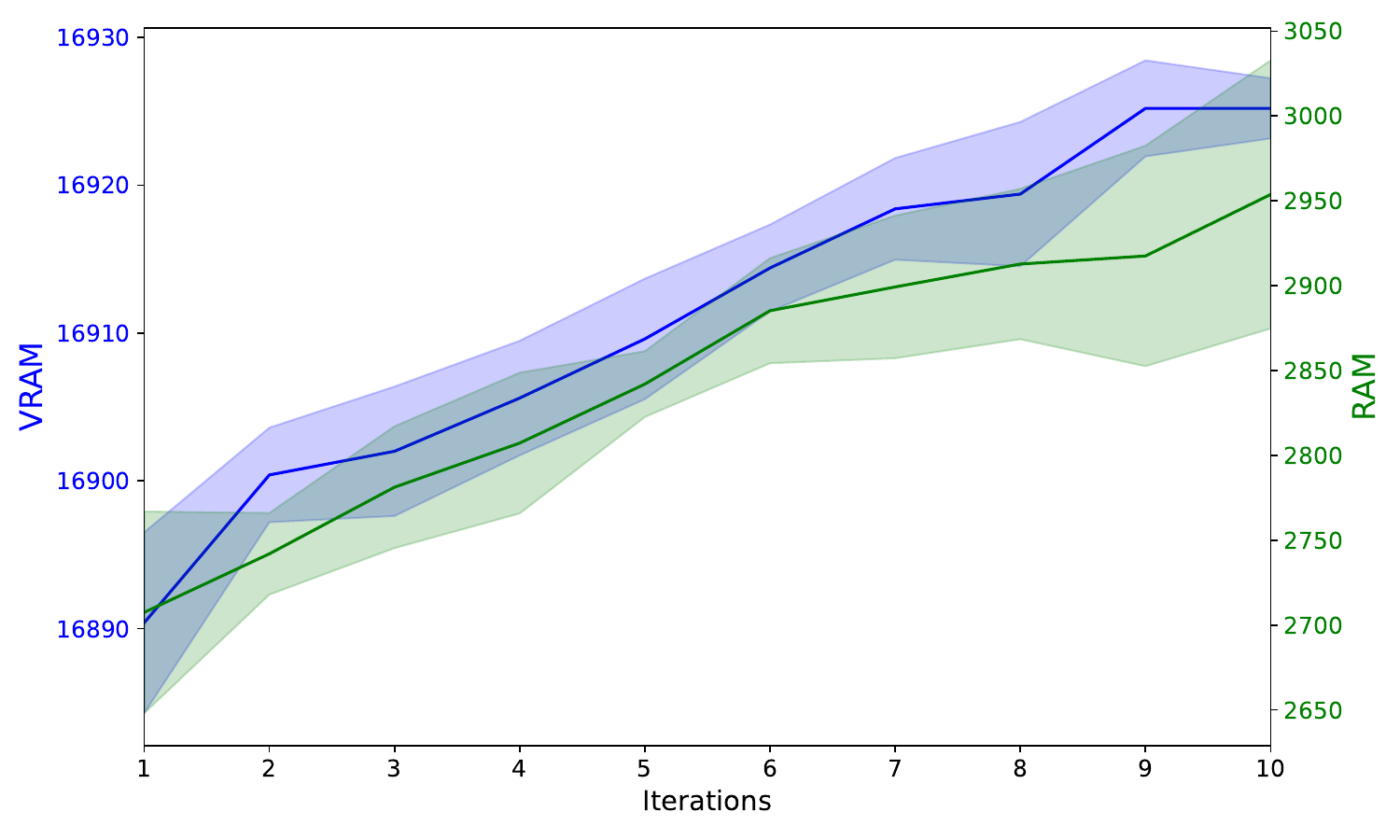}
    \vspace{-10pt}
    \caption{\small \textbf{Analysis on computational resources for iterative editing.}}
    \label{fig:comp_resource}
    \vspace{-10pt}
\end{figure}

\begin{figure*}[ht]
  \centering
  \includegraphics[width=0.9\textwidth]{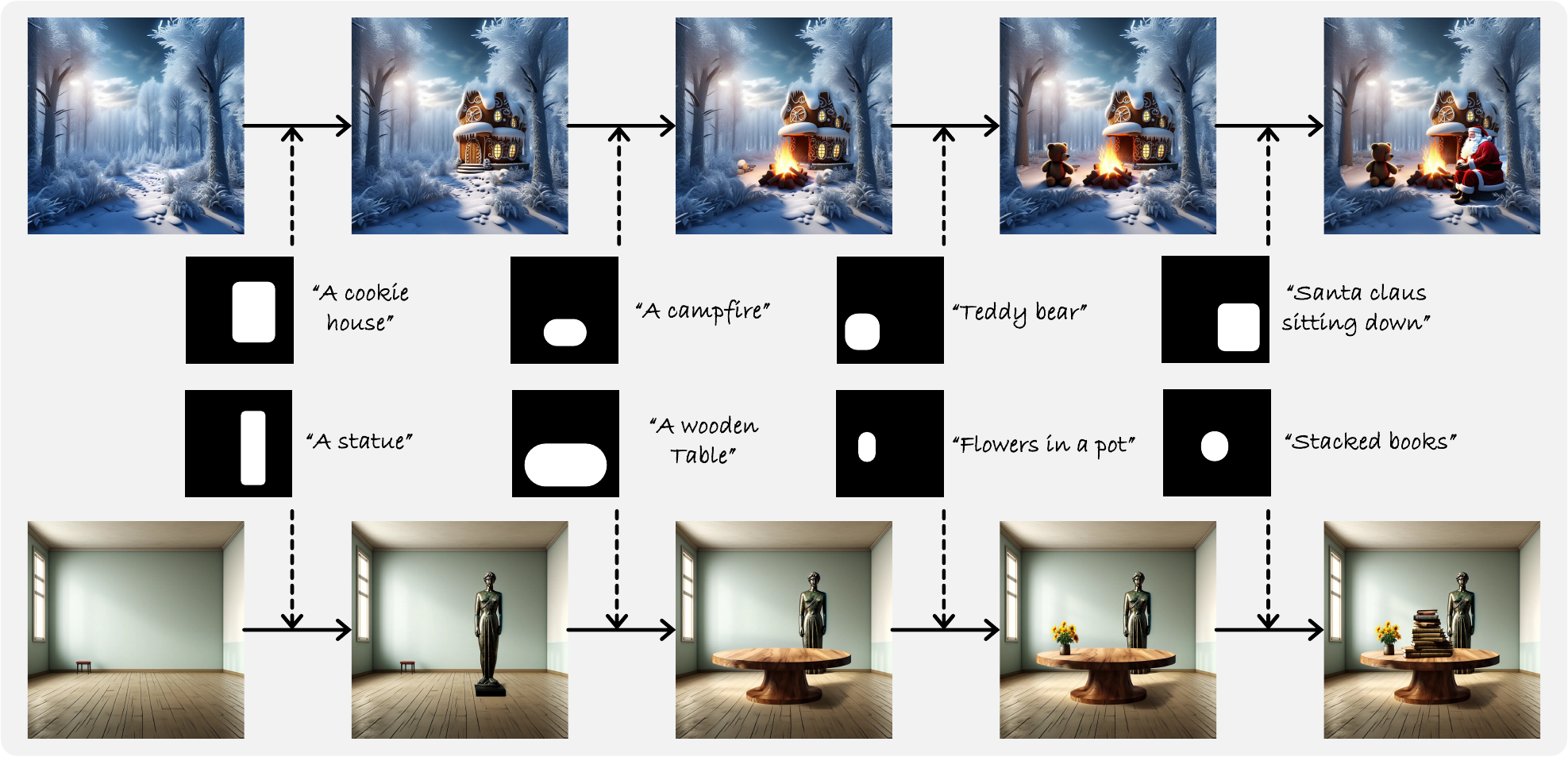}
  \vspace{-1mm}
  \caption{\textbf{Extensive multi-editing scenario.} Our framework enables sequential editing of multiple edits, more than just two or three times editing, meeting the user's need to edit extensively on generated images.}
  \label{fig:extensive_edit}
  \vspace{-1mm}
\end{figure*}

\section{Analysis on Computational Overhead}
Sequential editing multiple times, as in \cref{fig:extensive_edit,fig:appen_interactive}, can make the user achieve the intended images.
However, multiple editing with layer-wise memory requires additional computational cost, and we analyze computational resource utilization during iterative editing processes. 
We create a new dataset for this analysis following a similar generation protocol as Multi-Edit Bench and perform 5 independent trials of sequential edits up to 10 iterations, measuring both memory consumption and processing overhead.

Fig.~\ref{fig:comp_resource} illustrates the resource utilization patterns on a single NVIDIA RTX-A6000 GPU. 
Due to our layer-wise memory architecture, we observe a predictable linear increase in RAM usage from 2,653MB to 2,954MB across 10 iterations, representing an 11\% increment over the base memory footprint. 
This moderate increase is attributed to the storage of latent representations necessary for maintaining edit history and ensuring consistency across modifications.
Notably, the VRAM consumption shows remarkable efficiency, increasing marginally from 16,882MB to 16,925MB - a mere 0.2\% overhead over the initial usage.

\section{Perceptual Study}
\begin{table}[t]
    \centering
    \footnotesize
    \caption{\textbf{Human preference study results with recent foundational models~\cite{chen2023pixartalpha,podell2023sdxl}.} We evaluate our model with SD-XL and PixArt-alpha on various prompts regarding spatial relationships on alignment and overall image quality.}
    \vspace{-2mm}
    \resizebox{0.8\columnwidth}{!}{
    \begin{tabular}{@{}cccccc@{}}
    \toprule
     Method & Spatial Alignment & Overall Quality \\ 
       \cmidrule(lr){1-1} \cmidrule(lr){2-3} 
     SD-XL~\cite{li2023gligen} & 3.16 & 3.66 \\
     PixArt-alpha~\cite{avrahami2023blended} & 2.76 & 3.43 \\
     \midrule
     Ours & 3.57 & 3.47 \\
   \bottomrule
    \end{tabular}
    }
\label{tab:human}
\end{table}
We additionally compare complex scenarios with recent Text-to-Image models~\cite{podell2023sdxl, chen2023pixartalpha} and compare it in two aspects: (1) spatial alignment with the user's intention and (2) overall quality. 
The result is shown in \cref{tab:human}.

Our approach outperforms these latest models by more than 0.4 Likert scale in spatial alignment, showing the capability of synthesizing images while aligning well with a user's intention through an interactive editing process.
Furthermore, through multiple editing processes, we maintain the overall quality of the image (\ie, natural blending), achieving competitive results with recent models with a score of 3.47.
While this performs slightly lower than SD-XL, it shows improvement over the original PixArt-alpha model it builds upon.

\section{Dataset and Benchmark}
\label{sec:dataset}

In this section, we first showcase our result on other benchmarks for single-turn editing~\cite{editbench}.
Afterward, we discuss the limitations of existing datasets and benchmarks, particularly in the context of interactive image generation and sequential image editing. 
We present details of the benchmark proposed in \textbf{Sec.4} of the manuscript, which is designed to address these shortcomings by introducing scenarios tailored to evaluate spatial arrangement and semantic alignment in iterative editing tasks.

\subsection{Comparison on EditBench}

We evaluate our framework on EditBench~\cite{editbench} to assess its performance on single-turn image editing scenarios. 
As shown in \cref{tab:editbench}, our method achieves competitive results on EditBench's metrics, demonstrating CLIP Text-to-Image scores and R-Precision (Prec.) comparable to state-of-the-art methods like Blended Latent Diffusion (BLD) with SD-XL and HD-Painter.

For CLIP Text-to-Image (T2I) score, ours outperforms all the baselines of Blended Latent Diffusion (BLD) with SD-XL, HD-Painter, and SD3-ControlNet-Inpaint.
Also, ours outperforms Imagen-Editor~\cite{editbench} in CLIP T2I score, demonstrating the effectiveness.
Especially, SD3-Inpaint showcases competitive results to ours in single-turn edits, but they show lower performance compared to BLD or HD-Painter in multiple edits demonstrated in \textbf{Sec. 5} of the main paper.
Also, BLD and HD-Painter show lower performance on CLIP-Score in the result of \textbf{Sec. 5}.
This demonstrates that traditional methods like BLD, HD-Painter, and SD-3-ControlNet-Inpaint are quite effective for single edits. However, they struggle with maintaining consistency across multiple editing steps as they lack mechanisms for preserving editing history and ensuring cross-edit coherence.
This highlights a limitation of current benchmarks like EditBench that focus solely on single-turn editing. 

\begin{table}[t]
    \centering
    \footnotesize
    \caption{\textbf{Comparison of latest works on single-turn editing.} We evaluate our model with Blended Latent Diffusion (BLD) with SD-XL and HD-Painter on single inpainting on EditBench. Following EditBench, we evaluate the CLIP Text-to-Image (T2I) score and CLIP R-Precision (Prec.). IM denotes Imagen-Editor proposed in EditBench.~\cite{editbench}}
    \vspace{-2mm}
    \resizebox{0.85\columnwidth}{!}{
    \begin{tabular}{@{}ccccccc@{}}
    \toprule
     Training & Method & CLIP (T2I) & CLIP R-Prec. \\ 
    \cmidrule(lr){1-1} \cmidrule(lr){2-2} \cmidrule(lr){3-4}
    O & IM & 31.5 & \textbf{98.6}\\
    \midrule
     \multirow{4.5}{*}{X} &BLD & 29.84 & 70.83\\
     & HD-Painter & 31.44& 87.50  \\
     & SD3-Inpaint & 31.65 & 87.92\\
     \cmidrule(lr){2-4}
     &Ours & \textbf{31.69}& 90.42\\
   \bottomrule
    \end{tabular}
    }
\label{tab:editbench}
\end{table}

\subsection{Limitations of Existing Datasets}
\label{sec:limitations}

Existing datasets and benchmarks in image editing~\cite{i2ebench, editbench} or image synthesis~\cite{bakr2023hrs, feng2023layoutgpt} often fail to evaluate the complex tasks involved in interactive image generation adequately.
Most notably, they fail to assess how well-generated images align with specific prompts and spatial relationships in the editing or generation process.
To summarize, prior works have the following limitations:

\begin{itemize}
    \item \textbf{Lack of Interactive Generation Evaluation:} Current benchmarks do not provide an effective means to evaluate interactive generation scenarios where objects are introduced sequentially into a scene with precise control over spatial arrangements.
    \item \textbf{Lack of Semantic Alignment Evaluation:} Evaluating the semantic alignment between the generated image and the prompt is often reduced to general-purpose metrics such as the CLIP score or mean Average Precision (mAP) from object detection models~\cite{yolo}. These metrics are insufficient to measure how well the generated image aligns with the intended semantics of the prompt, especially in complex, layered scenarios.
    \item \textbf{Inadequacy for Mask Order-aware Arrangement Evaluation:} Existing datasets are not designed to assess spatial relationships and image ordering. They rarely focus on occlusions or specific arrangements of objects in depth-aware compositions, making it difficult to evaluate whether the edited image faithfully captures the intention.
\end{itemize}

\noindent Considering these limitations, a novel benchmark is required to evaluate both interactive generation and editing scenarios while ensuring strong alignment with the input prompts.

\subsection{Details of Proposed Benchmark}
\label{sec:benchmark}

We introduce a new benchmark for evaluating sequential image generation and editing interactively, focusing specifically on the limitations mentioned above.
This benchmark introduces novel evaluation metrics and scenarios that rigorously test the model’s ability to generate images aligned with spatial constraints (\ie, mask for inpainting) and semantic intent.

\paragraph{Design.}

The proposed benchmark is crafted to assess the performance of models in generating images under an interactive generation scenario with sequential iterative editing.
Specifically, this includes the following components:
\begin{itemize}
    \item \textbf{Mask-ordered Prompts and Masks:} Each image generation task involves sequential prompts and corresponding masks that define the region of interest (RoI) for each object. This simulates an interactive generation process where objects are introduced layer by layer.
    \item \textbf{High Occlusion Ratio:} The benchmark is designed to test mask order-aware generation, ensuring that the scenarios involve significant object occlusion, a critical factor in realistic editing.
    \item \textbf{Complex Backgrounds and Detailed Object Arrangements:} Each scene includes a detailed and complex background, as well as intricate arrangements of objects, requiring the model to manage both background consistency and precise object placement effectively.
\end{itemize}

\paragraph{Features.}
To evaluate both the spatial alignment and overall visual quality of generated images, our benchmark includes the following features:
\begin{itemize}
    \item \textbf{Evaluation of Spatial Alignment:} The benchmark introduces tasks that require the model to add objects in specified spatial arrangements while maintaining spatial relationships after editing.
    \item \textbf{Semantic \& Visual-alignment Evaluation Metrics:} We propose several evaluation metrics that measure the alignment quality between the generated image and the intended prompt.
\end{itemize}

\begin{figure}[t]
  \centering
  \scriptsize
    \centering
    \includegraphics[width=\linewidth]{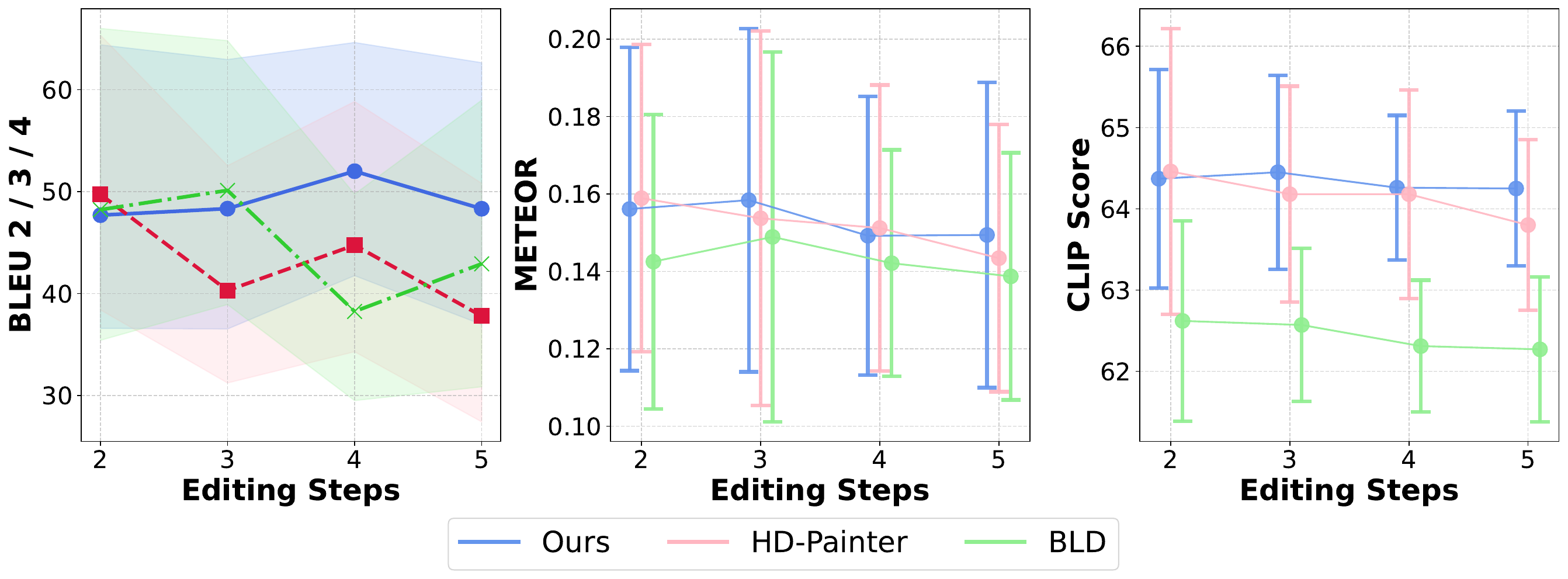
    } 
    \vspace{-3mm}
    \caption{{\textbf{Comparison of BLEU, METOER, and CLIP score on each step.}}}
    \label{fig:bleu_step}
    \vspace{-3mm}
\end{figure}

\subsection{Dataset Generation Process}
\label{sec:dataset_generation}

We generate the benchmark through a semi-automated process that ensures diversity in composition while maintaining a high degree of control over spatial relationships and occlusions.
The details for each step of the dataset generation process are as follows:

\textbf{Step 1: Decide on the number of layers (n):} Each image in the dataset consists of \( n \) layers, where \( n \) ranges between 3 and 6, including the background layer.

\textbf{Step 2: Select reference class from ImageNet-1K:} One object class is selected as the reference class from ImageNet-1K. This class serves as the anchor for the composition.

\textbf{Step 3: Select additional classes via GPT-4:} Using the GPT-4 API, \( n-1 \) additional classes are selected based on their natural compositional compatibility with the reference class. This ensures that the objects in the scene follow a coherent visual and semantic composition.

\textbf{Step 4: Generate random layouts (masks) for \( n \) classes:} For each of the \( n \) classes, random layouts are generated with constraints such as ``margin from the edges'' and the ``size of mask''. These constraints ensure the objects are well-distributed without excessive overlap or clutter.

\textbf{Step 5: Generate template-based captions:} Based on the center coordinates of each object mask, template-based captions are generated to describe the spatial relationships and contents of the scene. 
These templates are used to generate global captions for the entire scene and for individual layers regarding the spatial relations.

\textbf{Step 6: Generate global and layer-wise captions:} The global caption is generated to describe the entire scene, while individual layer-wise (\ie, editing steps) captions are generated for each object, ensuring that background details are excluded from the layer-wise descriptions with template-based captions through GPT API.

Through this approach, our dataset is designed to rigorously evaluate models' performance: capabilities in handling interactive generation scenarios, spatial alignments, and semantic accuracy within complex, mask order-aware environments.
As a result of this rigorous dataset construction, our benchmark evaluates editing performance across 2 to 5 steps, with distributions of 19\% (2-step), 18\% (3-step), 26\% (4-step), and 37\% (5-step), with average occlusion ratio of 18.53\% across the layers.

\subsection{Evaluation Details}
\label{sec:evaluation_details}

We evaluate each individual editing step of the edited image by cropping the generated image based on the masks provided for each step. 
This method allows for fine-grained evaluation of how well each individual object was added following its corresponding prompt and spatial arrangement.

\begin{figure}[t]
  \centering
  \vspace{-4mm}
  \scriptsize
    \includegraphics[width=\linewidth]{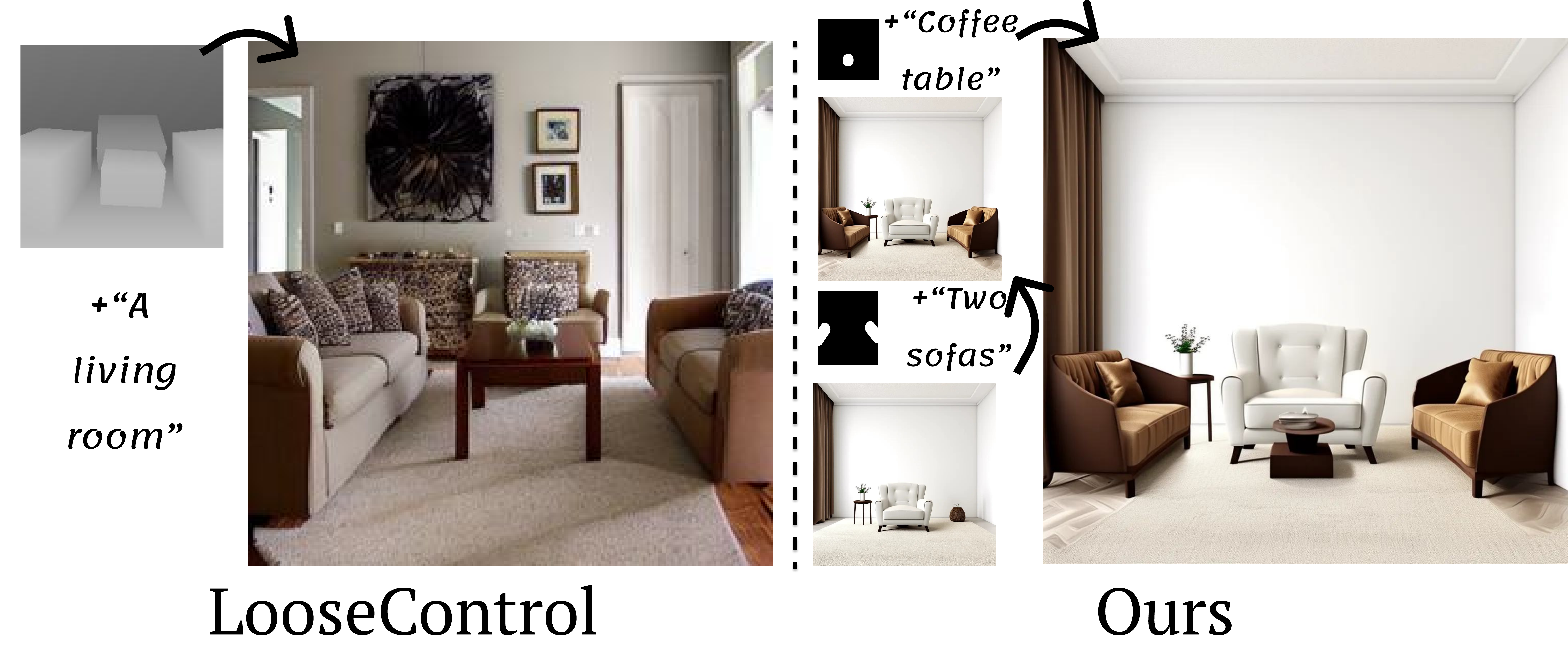}
    \vspace{-2mm}
    \caption{\textbf{Qualitative comparison on LooseControl.}}
    \label{fig:loose}
\end{figure}

\begin{figure}[t]
    \centering
    \captionof{table}{\textbf{Quantitative comparison on LooseControl.} $\dagger$ denotes Attribute Editing with cross-frame attention in LooseControl.}
    \vspace{-3mm}
    \setlength{\tabcolsep}{1.5pt}
    \resizebox{1.0\columnwidth}{!}{
    \begin{tabular}{lcccc}
    \toprule
      \multirow{2}{*}{Method} & \multicolumn{2}{c}{Semantic Align} & Visual Align \\ 
      \cmidrule(lr){2-3} \cmidrule(lr){4-4} 
       & BLEU-2/3/4$\uparrow$ & METEOR$\uparrow$ & CLIP\textsubscript{crop}$\uparrow$ \\
    \midrule
    LayoutGuidance & 36.44 / 26.13 / 18.85 & 0.1361 &  62.92 \\
    NoiseCollage  & 55.75 / 42.43 / 32.96 & 0.1402 & 64.01 \\
    \midrule
    LooseControl & 63.30 / 46.24 / 34.15 & 0.1373 & 63.13 \\
    LooseControl + Edit$\dagger$ &58.74 / 45.00 / 34.76 & 0.1359 & 62.32 \\
    \midrule
     Ours ($512\times512$) & 61.19 / 45.04 / 34.06 & 0.1465 & 64.28 \\
     Ours ($1024\times1024$) & \textbf{64.99} / \textbf{47.69} / \textbf{36.59} & \textbf{0.1513} & \textbf{64.29} \\
    \bottomrule
    \end{tabular}
    }
  \vspace{-4mm}
  \label{tab:loose}
\end{figure}

\paragraph{Cropped Image Evaluation}

All cropped images from the individual editing step's evaluation are resized to a resolution of $224\times224$ for evaluation.
This uniform resolution ensures that variations in image size do not introduce inconsistencies in the evaluation results.
The evaluation metrics used on the cropped images include the following metrics:
\begin{itemize}
    \item \textbf{CLIP Score:} We measure the similarity between each cropped image and its corresponding prompt. Since CLIP’s text encoder input is limited to 77 tokens and our prompt exceeds its length, CLIP score's expressiveness can be constrained~\cite{urbanek2024picture,zhang2024long}. Hence, we adopt the template \textit{``An image of \{CLASS\} in \{BACKGROUND\}''} to describe the local cropped region within the token limit.
    \item \textbf{LLaVa-based Alignment with BLEU and METEOR} For each cropped editing step's image, LLaVa generates captions based on the bounding box of each object. The alignment between these captions and the intended prompt is measured using BLEU and METEOR scores, ensuring the model accurately captures the intended semantic information for all editing steps.
\end{itemize}
\begin{figure*}[t]
  \centering
  \includegraphics[clip, width=0.75\textwidth]{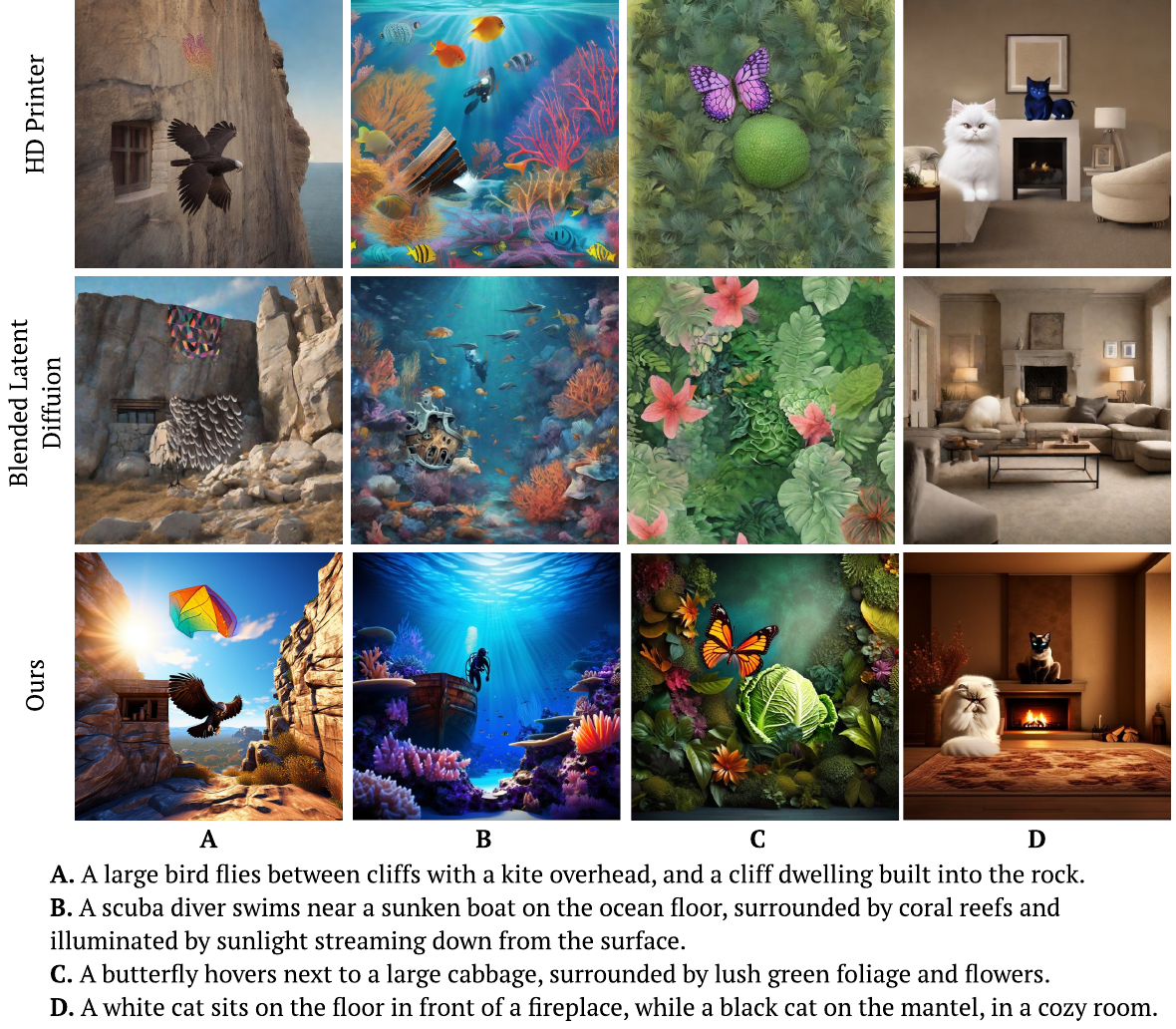}
  \caption{\textbf{Comparison with other latest editing approaches~\cite{avrahami2022blended,manukyan2023hd} with Multi-Edit Bench Dataset.} The approaches in the first two rows show results with baseline editing approaches. The background image is generated by our framework.}
  \label{fig:edit_quali_ds}
  \vspace{-1mm}
\end{figure*}

By evaluating each individual editing step, we ensure a comprehensive assessment of the model’s ability to edit holistically in a spatially aligned and semantically accurate manner.

\subsection{Effect of Editing Steps}
We conduct additional experiments on editing steps and present the results in Fig.~\ref{fig:bleu_step} with BLEU, METEOR, and CLIP scores. Our method maintains stable performance as steps increase, whereas BLD and HD-Painter exhibit a continuous decline in CLIP and METEOR after three steps, along with consistently lower BLEU. Overall, our method remains steady across all metrics as editing steps increase.

\subsection{Comparison with 3D-lifted Work}
We further compare our method with 3D-lifted approaches~\cite{loosecontrol,eldesokey2024build}.  
Since Build-A-Scene~\cite{eldesokey2024build} is unavailable, we evaluate against LooseControl~\cite{loosecontrol} and its 3D-Editing approach in Tab.~\ref{tab:loose} and Fig.~\ref{fig:loose}.  

For fair evaluation, we lift 2D boxes to 3D using pseudo-depth maps and project them back for appropriate mask usage.  
LooseControl outperforms in BLEU at $512\times512$ but lags in METEOR and CLIP, while our method surpasses across all metrics at $1024 \times 1024$ resolution.  

Additionally, we compare with LooseControl's attribute editing.  
In multi-step editing, LooseControl consistently underperforms across all metrics compared to our method.  

\section{Qualitative Results}

We provide extensive qualitative results demonstrating our framework's versatility in handling various image editing scenarios.
Fig.~\ref{fig:appen_interactive} demonstrates the capability of interactive image generation under diverse scenarios.
We also provide \cref{fig:extensive_edit} to demonstrate the effectiveness of image synthesis under extensive multi-editing scenarios.

As we tackle the challenge of multiple editing, we showcase the qualitative comparison on our proposed Multi-Edit Bench in Sec.~\ref{sec:qual_multieditbench}
In addition, we present more qualitative result on advanced editing (\ie, deleting the object behind the generated object in overlapped scenario.) under Sec.~\ref{sec:quali_improved}.
Furthermore, we show the application of our method in depth-order aware generation in Sec.~\ref{sec:quali_depth} by comparing with depth-aware approaches, denoting our model's possibility in order-aware generation empowered by Background Consistency Guidance (BCG) and Multi-Query Disentangled cross-attention (MQD), maintaining the overlapped object's shape and context even we add an additional object with high occlusion ratio.

\subsection{Comparison on Multi-Edit Bench}
\label{sec:qual_multieditbench}

We present the result on Multi-Edit Benchmark dataset.
Note that we utilized prompts inside our generated dataset.
Due to the lack of space, we omit the prompt as a short sentence inside the qualitative result.

\vspace{2mm}
\noindent\textbf{Qualitative Comparison.}
In \cref{fig:edit_quali_ds}, we present results on our Multi-Edit Bench, compared to other baselines of Blended Latent Diffusion (BLD)~\cite{avrahami2023blended} and HD-Painter~\cite{manukyan2023hd}.
HD-Painter shows a quality image, but as seen in column A, `kite' is not apparent in the image compared to the naturally blended kite in ours. 
For BLD, they fail to add objects in most examples, showing degraded image quality.
In contrast, ours show images that align with the given masks in the dataset.

\begin{figure*}[t]
  \centering
  \includegraphics[clip, width=0.65\textwidth]{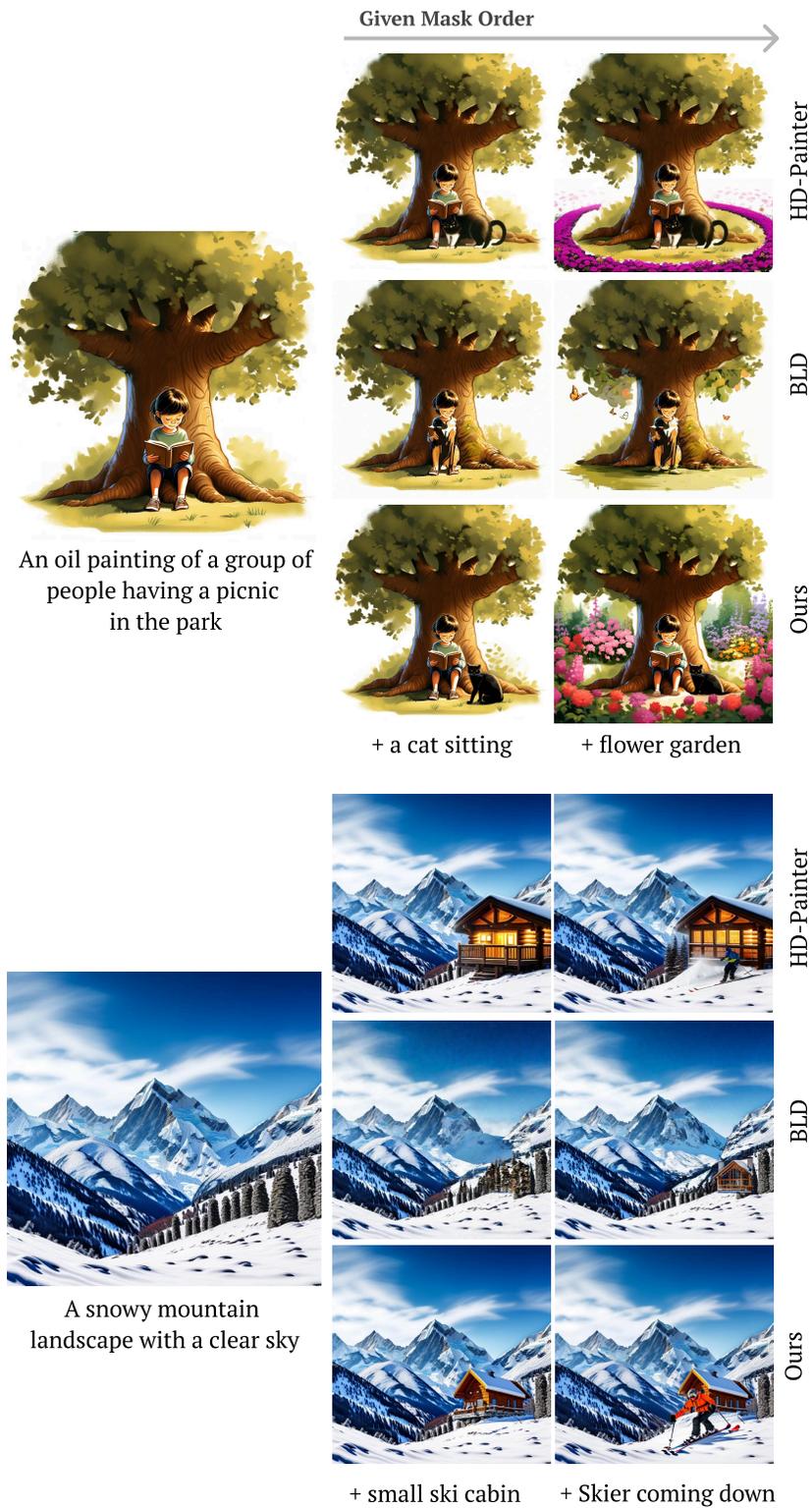}
  \caption{\textbf{Comparison with other latest editing approaches.} The approaches in the first two rows for each example show results with baseline editing approaches. The background image is generated by our framework.}
  \label{fig:edit_quali_arb1}
  \vspace{-1mm}
\end{figure*}

\vspace{2mm}
\noindent\textbf{Comparison under Real-world Interaction Scenarios.}
As we proposed Multi-Edit Bench to evaluate sequential editing scenarios, we additionally compare with arbitrary cases, as this work focuses on interactive generation. We gave arbitrary prompts and masks to look out for more interactive editing scenarios.
We designed prompts and masks arbitrarily but used the same prompts and masks for all the baseline models. We sampled 5 times and used the best-appearing sample for the qualitative comparison.
We showcase the comparison in \cref{fig:edit_quali_arb1} and \cref{fig:edit_quali_arb2}.

\subsection{Comparison on Improved Editing}\label{sec:quali_improved}
\begin{figure*}[t]
  \centering
  \includegraphics[trim={4mm 4mm 4mm 4mm}, clip, width=0.99\textwidth]{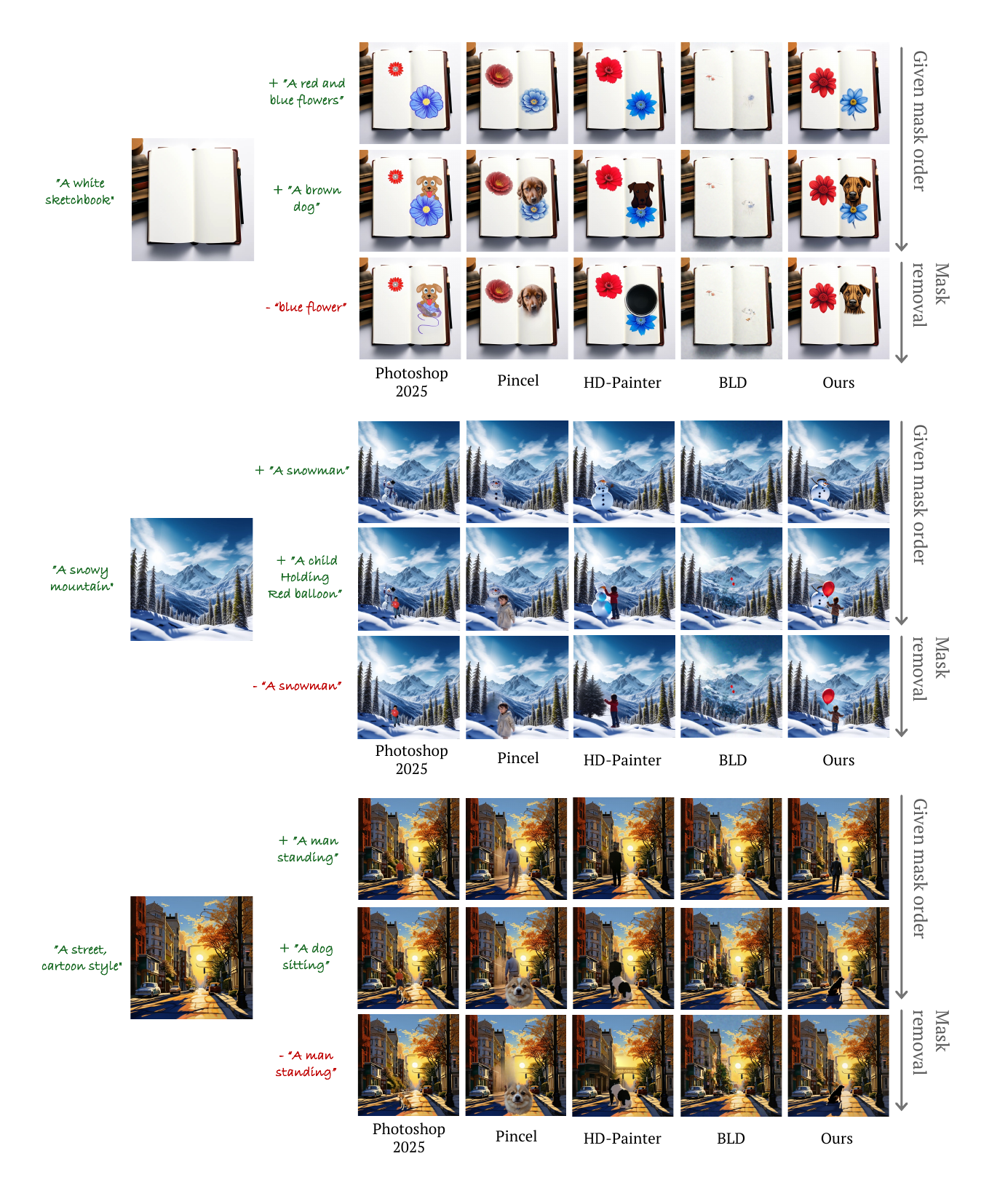}
  \vspace{-4mm}
  \caption{\textbf{Comparison under improved editing scenario.} Ours maintain the background well compared to other commercial products~\cite{photoshop,pincel} or baselines~\cite{avrahami2023blended, manukyan2023hd}.}
  \label{fig:improved_edit_quali_merged}
  \vspace{-1mm}
\end{figure*}

We present a qualitative comparison under an improved editing scenario in \cref{fig:improved_edit_quali_merged}.
To achieve improved editability, we utilize our method to delete the object behind the foreground object (\ie, previous mask order).
Other methods, including commercial products~\cite{photoshop, pincel} and baselines~\cite{avrahami2023blended,manukyan2023hd} show artifacts when removing the previously ordered object in the examples in \cref{fig:improved_edit_quali_merged}.
However, ours removes the previous object without any artifact, re-gaining the previous background through layer-wise memory.
Also, we maintain the foreground object's identity through MQD, describing our method's efficacy.

\subsection{Comparison with Depth-aware Approaches}\label{sec:quali_depth}

\begin{figure*}[t]
  \centering
  \includegraphics[clip, width=1.0\textwidth]{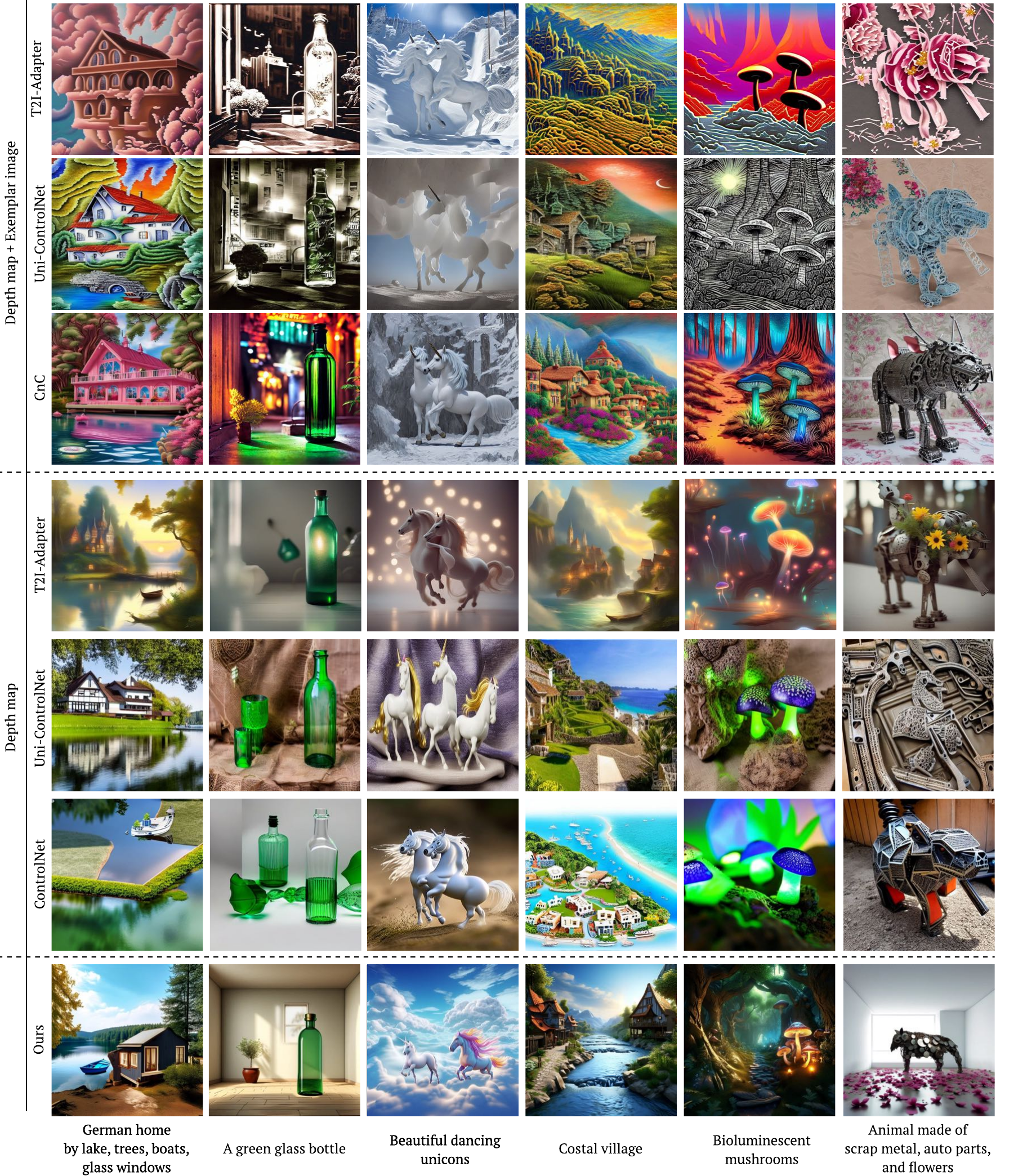}
  \vspace{-4mm}
  \caption{\textbf{Comparison with depth-aware text-to-image approaches.} The approaches in the first three rows utilize a depth map, exemplar image, and text prompt. The approaches in the next three rows get a depth map and text prompt. Our approach rivals the baseline approaches without using depth maps or exemplar images.}
  \label{fig:depth_quali}
  \vspace{-1mm}
\end{figure*}

We additionally compare our method with depth-aware models in \cref{fig:depth_quali}, as we can also generate order-aware images.
ControlNet~\cite{zhang2023adding}, T2I-Adapter~\cite{mou2023t2i}, or Uni-ControlNet~\cite{zhao2023uni} show artifacts, but ours show results following the user's intention, like the model which is trained from scratch~\cite{lee2024compose}.

\clearpage

\begin{figure*}[t]
  \centering
  \includegraphics[clip, width=0.65\textwidth]{figures/quali_interactive_2.pdf}
  \caption{\textbf{Comparison with other latest editing approaches.}The approaches in the first two rows for each example show results with baseline editing approaches. The background image is generated by our framework.}
  \label{fig:edit_quali_arb2}
  \vspace{-1mm}
\end{figure*}

\end{document}